\documentclass[lettersize,journal]{IEEEtran}
\usepackage{amsmath,amsfonts}
\usepackage{algorithmic}
\usepackage{algorithm}
\usepackage{array}
\usepackage[caption=false,font=normalsize,labelfont=sf,textfont=sf]{subfig}
\usepackage{textcomp}
\usepackage{stfloats}
\usepackage{url}
\usepackage{verbatim}
\usepackage{graphicx}
\usepackage{cite}
\begin{document}

\title{Accelerating Diffusion for SAR-to-Optical Image Translation via Adversarial Consistency Distillation}

\author{Xinyu Bai,~\IEEEmembership{Student Member,~IEEE,} Feng Xu,~\IEEEmembership{Senior Member,~IEEE}
\thanks{The authors are with the Key Laboratory for Information Science of Electromagnetic Waves (MoE), Fudan University, Shanghai 200433, China. (e-mail: fengxu@fudan.edu.cn)}}



\maketitle

\begin{abstract}
Synthetic Aperture Radar (SAR) provides all-weather, high-resolution imaging capabilities, but its unique imaging mechanism often requires expert interpretation, limiting its widespread applicability.  Translating SAR images into more easily recognizable optical images using diffusion models helps address this challenge. However, diffusion models suffer from high latency due to numerous iterative inferences, while Generative Adversarial Networks (GANs) can achieve image translation with just a single iteration but often at the cost of image quality. To overcome these issues, we propose a new training framework for SAR-to-optical image translation that combines the strengths of both approaches. Our method employs consistency distillation to reduce iterative inference steps and integrates adversarial learning to ensure image clarity and minimize color shifts. Additionally, our approach allows for a trade-off between quality and speed, providing flexibility based on application requirements.
We conducted experiments on SEN12 and GF3 datasets, performing quantitative evaluations using Peak Signal-to-Noise Ratio (PSNR), Structural Similarity Index (SSIM), and Frechet Inception Distance (FID), as well as calculating the inference latency. The results demonstrate that our approach significantly improves inference speed by 131 times while maintaining the visual quality of the generated images, thus offering a robust and efficient solution for SAR-to-optical image translation. 
\end{abstract}


\section{Introduction}

\IEEEPARstart{S}{ynthetic} Aperture Radar (SAR) represents a highly advanced remote sensing technology that achieves high-resolution imaging by integrating radar principles with synthetic aperture methods\cite{sun2017recent}. Despite its significant advantages, SAR image interpretation is challenging due to issues such as speckle noise, geometric distortions, and radiometric calibration, often requiring the expertise of trained professionals\cite{choi2019speckle}. 
In contrast, optical images, which use visible light or near-infrared bands, provide intuitive spectral and spatial information about ground objects, making them easily recognizable. However, optical images are vulnerable to cloud cover and atmospheric scattering, leading to information loss or blurring\cite{bai2023conditional}. 
To address these challenges and improve SAR image interpretation efficiency, a method has emerged to translate SAR images into optical images. This approach simplifies SAR image understanding for non-specialists and leverages the clear and intuitive characteristics of optical images to overcome some limitations of SAR. By combining SAR and optical images, it provides comprehensive and detailed information about ground objects, easing the interpretation process and expanding application fields. This method holds promise in environmental monitoring, natural disaster assessment, and urban planning, significantly enhancing the practical value and utilization efficiency of SAR images.

In the task of SAR-to-optical image translation, traditional methods heavily rely on advanced machine learning algorithms, such as Support Vector Machines (SVMs)\cite{fu2012research123} and other neural networks\cite{tiwari2022data, zhang2017approach}, along with complex feature extraction and engineering design, which makes the entire process costly and inefficient. 
Recently, significant advancements in the field of deep learning, particularly the emergence of Generative Adversarial Networks (GANs)\cite{goodfellow2020generative} and Denoising Diffusion Probabilistic Models (DDPMs)\cite{ho2020denoising}, have introduced novel solutions to this task. These deep learning models can autonomously learn high-level semantic features of images end-to-end, eliminating the need for manual feature design and extraction. Such technologies enable the translation of SAR images into more human-friendly and interpretable optical images in a more efficient and automated manner, significantly enhancing the efficiency of remote sensing data analysis and applications. GANs, through adversarial interaction training between generator and discriminator neural networks, can generate synthetic optical images consistent with real data distributions, producing corresponding optical images with just a single iteration.

In recent years, DDPMs have gradually become mainstream methods for image synthesis, surpassing traditional GAN-based models\cite{dhariwal2021diffusion,rombach2022high,cao2024survey}. Unlike GANs, which operate through adversarial training between a generator and a discriminator, DDPMs generate images by starting from random Gaussian noise and iteratively denoising through a sampler to reconstruct a clean target image. This noise diffusion and denoising-based generation method avoids the common mode collapse issue in GANs, providing more stable and higher-quality image synthesis results. DDPMs' training process maximizes the variational bound of the negative log-likelihood, making DDPMs advantageous in generation quality and model robustness\cite{croitoru2023diffusion}. Recently, Bai et al.\cite{bai2023conditional} applied conditional diffusion models to the task of translating SAR images to optical images(S2ODPM), using SAR images as conditional inputs in the diffusion model's training and inference process, which successfully generated optical remote sensing images with clearer boundaries and higher fidelity details. The key to this work is incorporating the original SAR image data as prior conditions into the DDPM generation process, enabling the model to effectively utilize the structural and textural information in SAR images, thereby guiding the diffusion model to generate optical image representations that better match natural scenes.
\begin{figure}[htbp]
    \label{fig1}
    \centering
    \begin{tabular}{@{\hspace{1pt}}c@{\hspace{1pt}}c@{\hspace{1pt}}c@{\hspace{1pt}}c@{\hspace{1pt}}c}
    \includegraphics[width=0.095\textwidth]{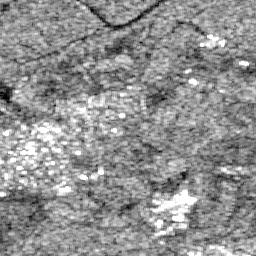}&
    \includegraphics[width=0.095\textwidth]{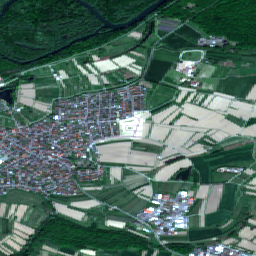}&
    \includegraphics[width=0.095\textwidth]{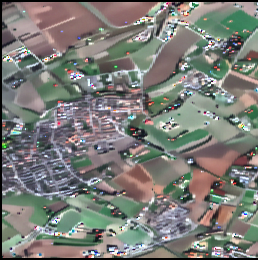}&
    \includegraphics[width=0.095\textwidth]{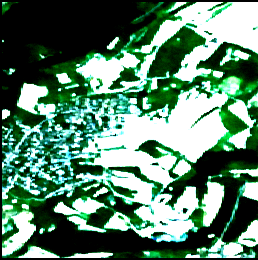} &
    \includegraphics[width=0.095\textwidth]{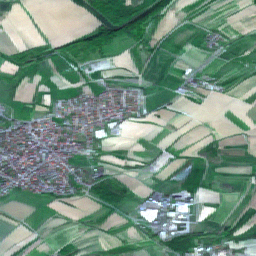}

\\
    \includegraphics[width=0.095\textwidth]{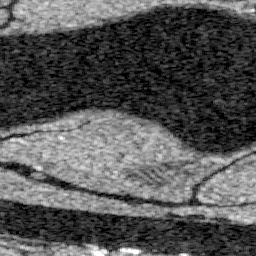}&
    \includegraphics[width=0.095\textwidth]{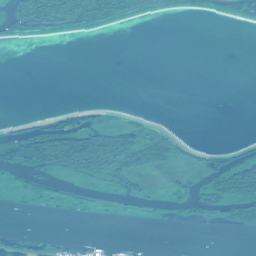}&
    \includegraphics[width=0.095\textwidth]{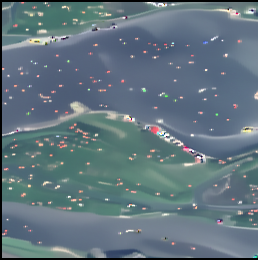}&
    \includegraphics[width=0.095\textwidth]{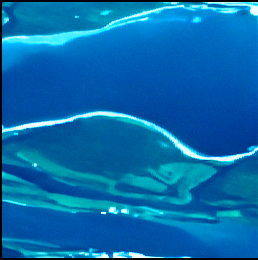}&
    \includegraphics[width=0.095\textwidth]{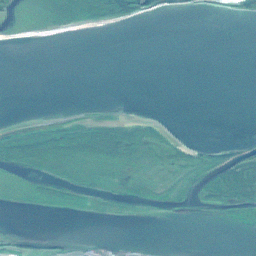}

\\
    \includegraphics[width=0.095\textwidth]{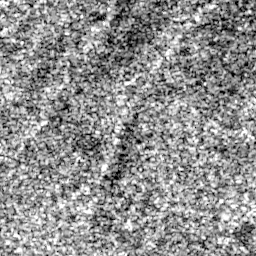}&
    \includegraphics[width=0.095\textwidth]{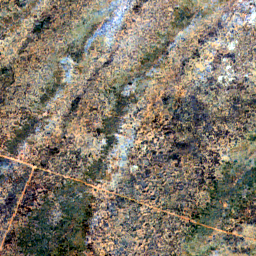}&
    \includegraphics[width=0.095\textwidth]{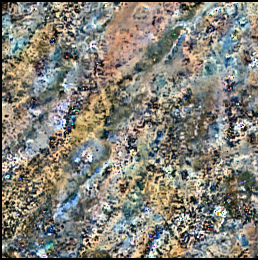}&
    \includegraphics[width=0.095\textwidth]{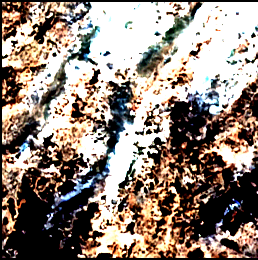} &
    \includegraphics[width=0.095\textwidth]{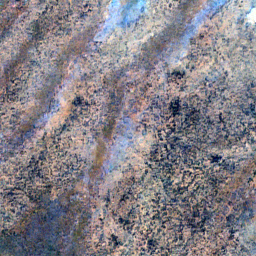}
\\
    \includegraphics[width=0.095\textwidth]{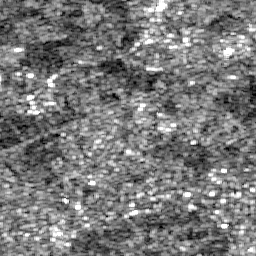}&
    \includegraphics[width=0.095\textwidth]{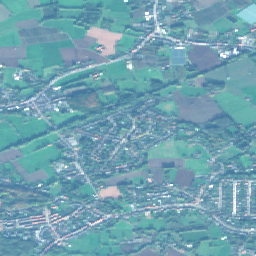}&
    \includegraphics[width=0.095\textwidth]{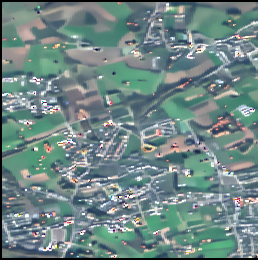}&
    \includegraphics[width=0.095\textwidth]{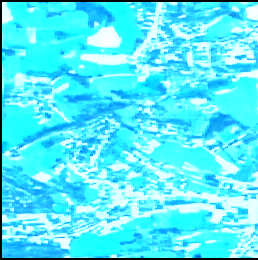} &
    \includegraphics[width=0.095\textwidth]{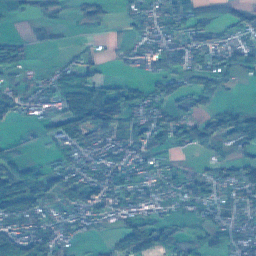}
\\

\ (a) & \ (b) & \ (c) & \ (d) & \ (e) \\
    \end{tabular}
    \caption{Comparison of different sampling methods for SAR-to-optical image translation. Columns from left to right: (a) SAR input images, (b) ground truth (GT) optical images, (c) results using the DDIM sampler, (d) results using the DPM++ sampler, and (e) results using the original sampler. The original method exhibits color shift, while the accelerated DDIM and DPM++ samplers exacerbate this issue, producing images that are often blurry and exhibit noticeable artifacts. This highlights the need for improved sampling methods to enhance image quality.}
\end{figure}

However, on the one hand, the iterative inference process of current diffusion models applied to SAR-to-optical image translation requires numerous sampling steps, resulting in slow inference speeds. On the other hand, these models also suffer from issues with color shifts. Although some studies have proposed fast samplers, such as DDIM\cite{song2020denoising} and DPM++\cite{lu2022dpm}, these training-free acceleration methods often result in blurry images with noticeable artifacts, exacerbating the existing color shift issue, as shown in Fig. \ref{fig1}. In contrast, GANs can achieve the translation process through a single iteration due to their training approach, which combines generator and discriminator dynamics.
To address these issues, inspired by Consistency Models (CM)\cite{song2023consistency}, we propose a DDPM Consistency Distillation method for SAR-to-optical image translation, incorporating an adversarial learning mechanism. The consistency distillation method ensures that the generated images are precise in detail and structure by maintaining consistency constraints between the source and target images. Meanwhile, through the interaction of the generator and discriminator, the adversarial learning mechanism further enhances image quality, making the generated optical images clearer and richer in detail.
By integrating the speed and training methodology of GANs into our framework, we combine the rapid translation capabilities of GANs with the superior image quality associated with diffusion models. Our experimental results demonstrate that this method effectively improves the quality and detail fidelity of generated images while accelerating the sampling process, overcoming the common issues of blurriness and artifacts in traditional fast sampling methods. These improvements significantly enhance the real-time application potential of SAR-to-optical image translation models, providing a more efficient and reliable remote sensing data analysis solution.

In summary, the main contributions of this paper are threefold: (1) We are the first to introduce the consistency distillation method to the SAR-to-optical image translation task. This method significantly accelerates the translation process by reducing the number of iterative inference steps while maintaining the high quality of the generated optical images. Consistency distillation also allows for a flexible trade-off between the number of iterations and image quality. (2) We designed a novel discriminator that integrates adversarial learning into the distillation process, reducing color shifts and enhancing the clarity of the generated optical images. (3) We conducted extensive experiments on the SEN12 and GF3 datasets, providing quantitative and qualitative evaluations of our model's performance. Our results show that this method effectively reduces the number of iterative inference steps while addressing common issues such as image blurriness and artifacts found in traditional fast sampling methods, demonstrating superior performance compared to GAN-based models.

\section{Related Work}
\subsection{SAR-to-Optical Image Translation}
In the field of remote sensing, interpreting SAR images is both essential and complex, driving a growing interest in transforming these images into more easily interpretable optical counterparts. Traditional methods have focused on coloring SAR images to highlight different targets, but they often fall short of accurately depicting actual ground scenes, leading to misinterpretations and inaccuracies\cite{deng2008colorization,uhlmann2013integrating}.

With the advancement of deep learning, techniques for translating SAR to optical images using deep learning have gradually emerged. These new methods aim to provide more accurate and contextually relevant representations of ground scenes, significantly improving the interpretability and utility of SAR data.
Merkle et al.\cite{fuentes2019sar} achieved the first alignment of remote sensing images and attempted to convert SAR images into optical images, demonstrating the significant potential of deep learning in the SAR-to-optical image translation field. The advent of GANs further stimulated research in this area, offering promising results in image-to-image translation tasks. 
Darbaghshahi et al.\cite{darbaghshahi2021cloud} utilized two types of GANs to convert SAR to optical images and remove cloud cover, introducing a dilated residual initial block in the generator. 
Yang et al.\cite{yang2022sar} developed ICGAN, which integrates multi-level features to enhance optical image contours. 
Fu et al.\cite{fu2021reciprocal} further introduced multi-scale discriminators to the GAN framework to better synthesize optical images from SAR data, addressing some of the limitations of previous methods.
CFRWD-GAN\cite{wei2023cfrwd} combines cross-fusion reasoning and wavelet decomposition to reduce speckle noise and recover high-frequency details in SAR images, thereby enhancing the translation quality to optical images.
Despite these advancements, GANs often face training difficulties and blurred boundaries. 

Recently, Bai et al.\cite{bai2023conditional} successfully applied diffusion models to the SAR-to-optical image translation task. They proposed a conditional diffusion model that incorporates SAR images as conditions into the training and inference process of the diffusion model, effectively overcoming the issues associated with GANs. 
However, this approach requires nearly a thousand iterative inference steps to obtain clear optical images, significantly hindering its practical application. To address this issue\cite{bai2023conditional}, we propose an adversarial consistency distillation method to accelerate the sampling process, considerably reducing the number of iterative inference steps without compromising image quality. This method combines the strengths of GANs and diffusion models, offering a more efficient and practical solution for SAR-to-optical image translation.

\subsection{Diffusion Models}
Diffusion models are a recently proposed advanced generative model that outperforms GANs in many computer vision tasks. These models can generate high-quality images from noise. Over recent years, diffusion models have rapidly evolved into a powerful generative modeling approach. 
Ho et al.\cite{ho2020denoising} introduced the Denoising Diffusion Probabilistic Model (DDPM), successfully applying diffusion models to the field of image generation. DDPMs have shown impressive performance in unconditional image generation and have made significant progress in various conditional generation tasks.
Building on the foundation of DDPM, numerous researchers have expanded and improved the model. 
Dhariwal et al.\cite{dhariwal2021diffusion} enhanced the model architecture and proposed a classifier guidance method, which utilizes gradients from a pre-trained classifier to guide the generation of target class images. 
Nichol and Dhariwal\cite{nichol2021improved} further introduced the Improved DDPM, which alters the noise schedule and model architecture to enhance the quality and diversity of generated samples. 
Additionally, Song et al.\cite{songscore} proposed a framework known as Score-based Generative Models, which trains a score function to guide the sampling process, significantly improving generative efficiency.
Ho et al.\cite{ho2022classifier} introduced Classifier-Free Guidance (CFG), a technique used in diffusion models to enhance the quality and control of generated images without needing a separate classifier model. CFG conditions the diffusion process on input attributes during training and adjusts the influence of this conditioning during sampling, thereby guiding the model to produce images that better adhere to the desired characteristics. 

Compared to GANs, diffusion models offer advantages in diversity, training stability, and scalability.
Diffusion models have been applied in numerous areas beyond image generation. For example, they have been used in text-to-3D generation \cite{pooledreamfusion}, personalization of text-to-image models\cite{ruiz2023dreambooth}, and other domains such as 3D representations, video generation, and motion synthesis\cite{yang2023diffusion}.
In this paper, we utilize a method that employs SAR images as conditional inputs to guide the generation process. By leveraging diffusion models' superior image generation capabilities, we can generate high-quality optical images, effectively achieving the translation from SAR to optical images. Our approach incorporates SAR images as conditions, utilizing diffusion models to produce high-quality optical images, thereby facilitating efficient SAR-to-optical image translation.

\subsection{Fast Sampling of Diffusion Models}
While diffusion models have achieved remarkable success in generating high-quality images, their slow sampling speed remains a bottleneck, hindering real-time applications. Current approaches to accelerating diffusion model sampling are mainly divided into training-free and training-based methods. Training-free acceleration primarily relies on various numerical sampling methods. 
For instance, Song et al.\cite{song2020denoising} proposed the DDIM sampler, which implements a deterministic reverse process instead of the traditional stochastic methods used in diffusion models, thereby enhancing sampling efficiency and flexibility. 
More recently, the DPM Solver++\cite{lu2022dpm} was introduced as a high-order solver utilizing a novel implicit midpoint method to improve sampling accuracy and stability. Additionally, this algorithm incorporates adaptive step size control, dynamically adjusting the step size based on data complexity to maintain high performance across different data and noise levels.

Training-based methods mainly leverage knowledge distillation techniques, such as progressive distillation\cite{salimans2022progressive} and guided distillation\cite{meng2023distillation}.
Song et al.\cite{song2023consistency} introduced the consistency model, which learns a direct mapping from noise to data by enforcing consistency regularization along the diffusion trajectory. This approach enables one-step generation while allowing multi-step sampling at the cost of some quality. 
However, training-free methods cannot be directly applied to the task of SAR-to-optical image translation using diffusion models, and they still require nearly a hundred iterations for inference, as shown in Fig.\ref{fig1}. On the other hand, current training-based methods often result in blurry images at low step counts. To address these challenges, we propose an adversarial consistency distillation method to generate clearer optical images with fewer iterations.

\section{Method}
We aim to translate SAR images into high-quality optical images with as few sampling steps as possible. The advantage of GANs lies in their ability to naturally achieve fast generation (requiring only one inference step), as they are trained to generate images in a single step through adversarial training with a discriminator. However, GANs often suffer from artifacts and unclear boundaries in the generated optical images. On the other hand, DDPMs, with their iterative inference process, can address these issues but at the cost of increased inference delay. Consistency distillation can reduce the number of required inference steps but still typically requires 100 or more steps.
Therefore, our method combines the strengths of both GANs and DDPMs. It further reduces the number of iterative inference steps while maintaining, or even surpassing, the image quality of the original DDPM model. By leveraging the rapid generation capabilities of GANs and the high-quality output of DDPMs, our approach effectively balances speed and performance in translating SAR images to optical images.
\begin{figure}[!t]
\centering
\includegraphics[width=0.46\textwidth]{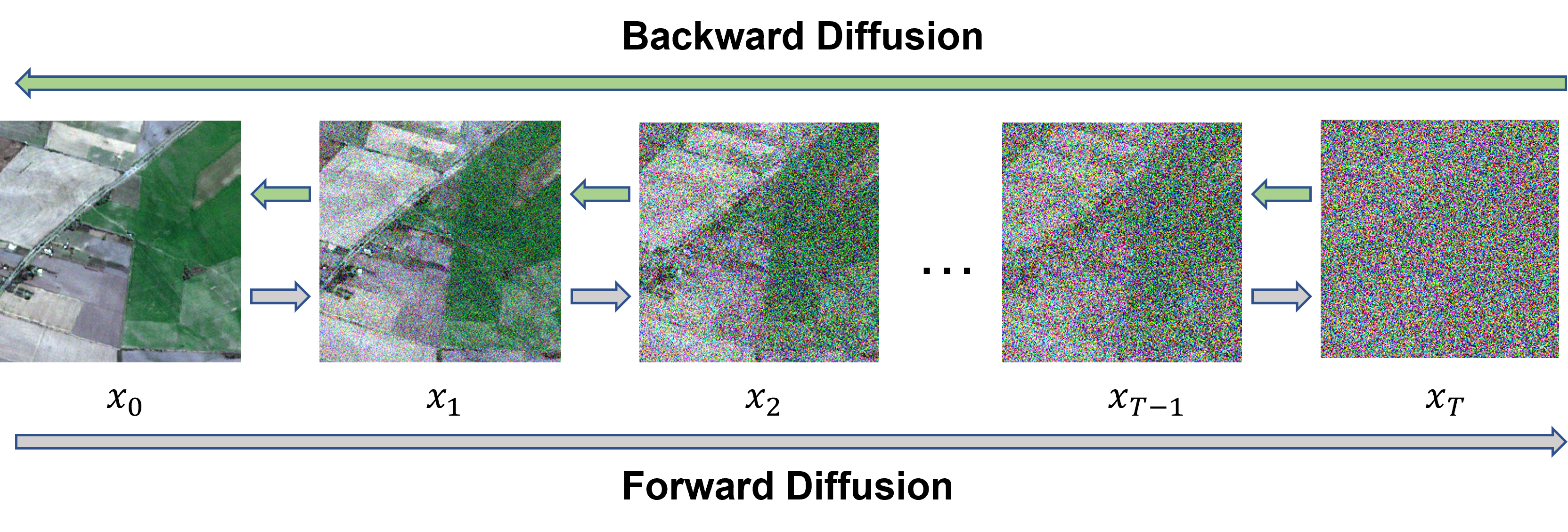}
\caption{Illustration of the diffusion process: the original optical image is progressively corrupted with Gaussian noise in the forward process and then denoised in the reverse process to reconstruct the original image. The arrows indicate the forward (right) and backward (left) processes.}
\label{diff}
\end{figure}
\subsection{Preliminaries}
\subsubsection{Diffusion Models}
Denoising Diffusion Probabilistic Models (DDPMs) work by iteratively removing noise from an initial pure Gaussian noise sample, thereby progressively revealing a clean data sample, as illustrated in Fig \ref{diff}.
Following the notations in \cite{songscore}, we briefly introduce the process.

The process begins with a predefined forward process \({x_t}\), indexed by a continuous time variable \( t \), during which Gaussian perturbations are progressively added to the data. This forward process can be modeled using a widely-adopted Stochastic Differential Equation (SDE)\cite{karras2022elucidating}:
\begin{equation}
\label{eq:1}
\mathrm{d} \boldsymbol{x}_{t} = \mu(t) \boldsymbol{x}_{t} \mathrm{~d} t + \nu(t) \mathrm{d} \boldsymbol{w}_{t} 
\end{equation}
where $\boldsymbol{w}_{t}$ represents standard Brownian motion, and $ \mu(t) $ and  $ \nu(t) $ are drift and diffusion coefficients, respectively. The marginal distribution of \( x_t \) along the forward process is denoted as \( p_t(x_t) \). This SDE gradually perturbs the empirical data distribution \( p_0(x) = p_{\text{data}}(x) \) towards the prior distribution \( p_t(x) \approx \pi(x) \), where \( \pi(x) \) is a tractable Gaussian distribution. Song et al.\cite{songscore} demonstrated that there exists an Ordinary Differential Equation (ODE) known as the Probability Flow (PF) ODE, whose trajectories maintain the same marginal probability densities \(\{p_t(x)\}_{t \in [0,T]}\) as the forward SDE:
\begin{equation}
\label{eq:2}
 \frac{\mathrm{d} \boldsymbol{x}_{t}}{\mathrm{~d} t} = \mu(t) \boldsymbol{x}_{t} - \frac{1}{2} \nu(t)^2 \nabla_{\boldsymbol{x}} \log p_t(\boldsymbol{x}_t)
\end{equation}
In practice, the ODE is solved along the trajectory defined by Equation \ref{eq:1}. This requires computing the score function\cite{hyvarinen2009estimation,song2019generative}, which is the gradient of the log-density with respect to the sample \( x \) and the noise level \( \sigma \) at each step. We can approximate this score function using a neural network parameterized by weights \( \theta \), trained for the denoising task. Through score matching, the true score in Equation \ref{eq:2} is approximated by the learned score model \( s_\theta(x, t) \approx \nabla_x \log p_t(x) \):
\begin{equation}
\label{eq:3}
\frac{\mathrm{d} \tilde{\boldsymbol{x}}_{t}}{\mathrm{~d} t} = \mu(t) \tilde{\boldsymbol{x}}_{t} - \frac{1}{2} \nu(t)^2 s_\theta(\tilde{\boldsymbol{x}}_{t}, t)
\end{equation}
Samples are then generated by solving the empirical PF ODE from \( T \) to 0. Efficient numerical solvers or pre-existing ODE solvers can be directly applied to approximate these solutions\cite{lu2022dpm1}.

In most applications\cite{choi2021ilvr, saharia2022palette}, each data sample \( x \) is associated with a condition \( c \). The output is controlled during generation by selecting the condition \( c \). In practice, this is achieved by training the denoising network to accept \( c \) as an additional conditional input.

\subsubsection{Consistency Distillation}
To tackle the challenge of generating samples for solving Equation \ref{eq:3}, which necessitates numerous neural network evaluations, Song et al.\cite{song2023consistency} proposed the consistency distillation training method. This approach achieves sample generation in a few steps by mapping any point on the PF ODE trajectory directly to its initial state:
\begin{equation}
f(x_t, t) = x_0, \quad t \in [0, T-1]
\end{equation}
This mapping is equivalent to maintaining the following self-consistency condition:
\begin{equation}
 f(x_t, t) = f(x_{t^\prime}, t^\prime), \quad t, t^\prime \in [0, T-1]
\end{equation}
To estimate the consistency function \( f \), they construct a parameterized model \( f_\theta \) and ensure its consistency properties through self-consistency. Typically, \( f \) can be extracted from a pre-trained diffusion model \( F_\theta \) and parameterized as:
\begin{equation}
\boldsymbol{f}_{\theta}(\boldsymbol{z}_t, \boldsymbol{c}, t) = c_{\text{skip}}(t) \boldsymbol{z}_t + c_{\text{out}}(t) F_{\theta}(\boldsymbol{x}_t, t)
\end{equation}
Here, $c_{\text{out}}(t)$ and $c_{\text{skip}}(t)$ are functions of time, where $c_{\text{out}}(t)$ gradually approaches 0 and $c_{\text{skip}}(t)$ gradually approaches 1 as time progresses.

To train the consistency model, a consistency distillation objective is defined by minimizing the following equation:
\begin{equation}
\begin{array}{l}
\mathcal{L}\left(\theta, \theta^{-} ; \phi\right) \\
\quad:=\mathbb{E}\left[\lambda\left(t_{n}\right)\left\|\boldsymbol{f}_{\theta}\left(\boldsymbol{x}_{t_{n+1}}, t_{n+1}\right)-\boldsymbol{f}_{\theta^{-}}\left(\hat{\boldsymbol{x}}_{t_{n}}^{\phi}, t_{n}\right)\right\|_{2}^{2}\right]
\end{array}
\end{equation}
where \( n \) is uniformly distributed over a given range and \( \lambda \) is a positive weighting function. Samples can be drawn using SDE, and the update function of a one-step ODE solver applied to the empirical PF ODE is used to represent them\cite{song2023consistency}. Additionally, to stabilize the training process, Song et al.\cite{song2023consistency} introduced \( \theta^{-} \) and updated it using an Exponential Moving Average (EMA) strategy.

By employing this consistency distillation approach, the model is able to generate samples efficiently, ensuring the consistency properties are maintained throughout the process. This training methodology significantly enhances the capability to produce high-quality images with fewer computational resources.
\begin{figure}[!t]
\centering
\includegraphics[width=0.5\textwidth]{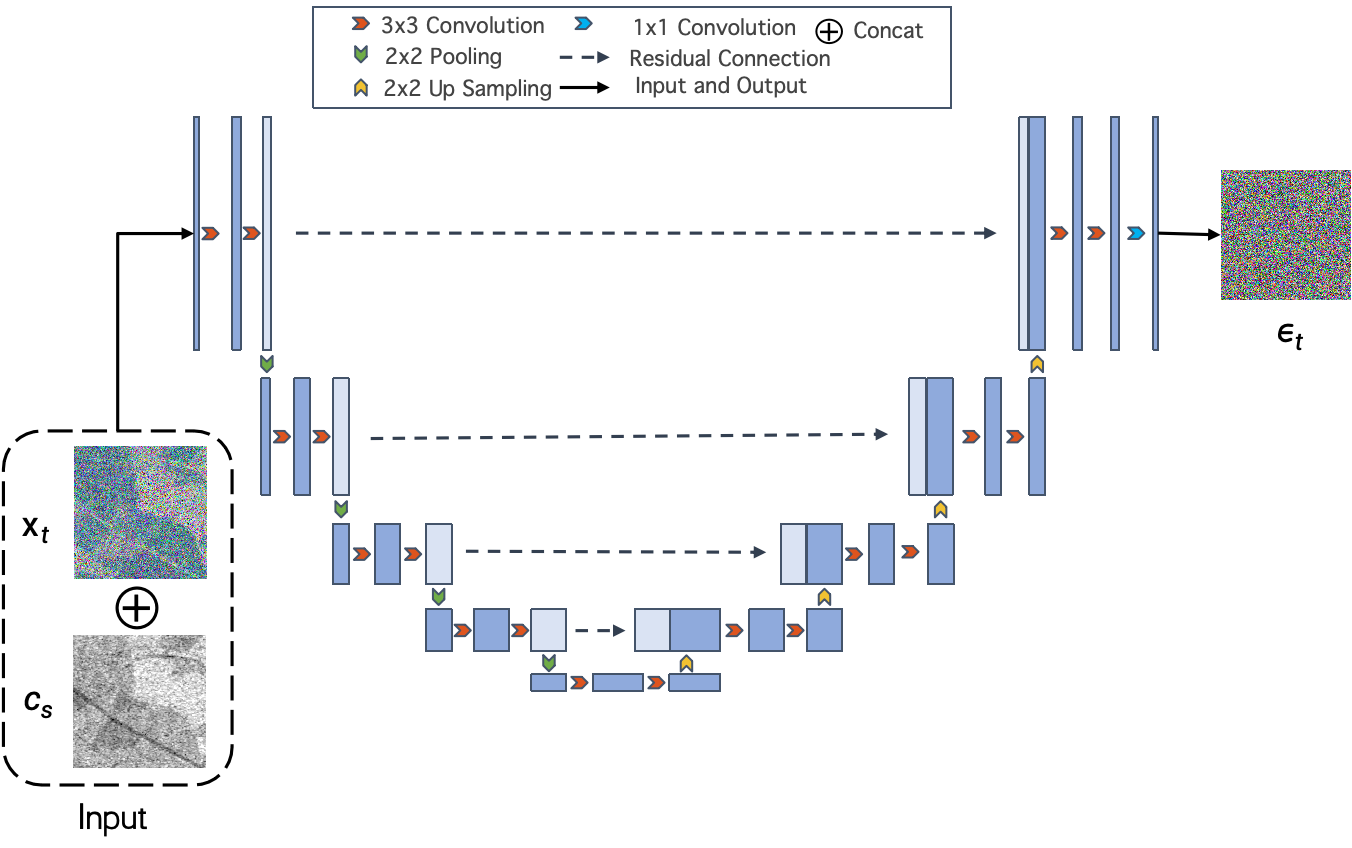}
\caption{The architecture of the teacher and student models based on the U-net network for noise prediction, used in the forward and backward diffusion processes for SAR-to-optical image translation.}
\label{unet}
\end{figure}
\subsection{Model Architecture}
Our teacher and student models follow the architecture proposed by Bai et al.\cite{bai2023conditional}, which utilizes a U-net network\cite{ronneberger2015u} primarily designed for noise prediction, as illustrated in Fig.\ref{unet}. These models are based on the original DDPM model with T diffusion steps, leveraging SAR images to guide the inference process. Specifically, in the forward diffusion process, an optical image \( x_t \) with added Gaussian noise is used as input, concatenated with the corresponding SAR image. During training, the model learns to predict the noise added at each diffusion step, conditioned on the SAR image.

In the backward generation process, the model begins with a Gaussian noise image \( x_T \) and iteratively denoises it step-by-step, conditioned on the SAR image, until \( x_T \) is transformed into a clear corresponding optical image over T steps. Notably, the SAR image condition at each diffusion step remains noise-free, providing the model with precise and consistent guidance. This approach ensures that the model receives accurate and stable inputs throughout the denoising process, thus improving the quality of the generated optical images\cite{bai2023conditional}.

\subsection{Discriminator Design}
Unlike traditional methods\cite{sauer2023adversarial}, our discriminator uses a simplified U-net network, retaining only the downsampling and intermediate layers and adding a mapping head. We believe that by using only the downsampling part of a well-trained U-net for SAR-to-optical image translation, we can effectively extract the critical features of the image while avoiding the computational burden associated with upsampling. This approach enhances overall training efficiency and stability. The extracted features are processed through the intermediate layers of the U-net and then mapped using a lightweight mapping head to assess the quality and authenticity of the images. This design not only simplifies the network structure but also ensures the extraction and processing of critical features, thereby improving the model's overall performance and training speed.

Specifically, our discriminator takes the SAR image as a condition concatenated with the generated or the real optical image. The combined input is then fed into the downsampling layers of the U-net. These layers are responsible for extracting multi-scale features from the input images, capturing essential details and structural information\cite{ronneberger2015u}.
The intermediate layers of the U-net further process these extracted features, refining and enhancing them. Finally, the processed features are passed through a lightweight mapping head, which assesses the quality and authenticity of the images. The mapping head outputs a scalar value indicating whether the input image is real or generated.

This design simplifies the network structure while ensuring effective feature extraction and processing. By concentrating on downsampling and intermediate layer processing, our discriminator efficiently evaluates the realism of the generated images, contributing to the adversarial learning process and ultimately enhancing the quality of SAR-to-optical image translations.
\begin{figure}[!t]
\centering
\includegraphics[width=0.45\textwidth]{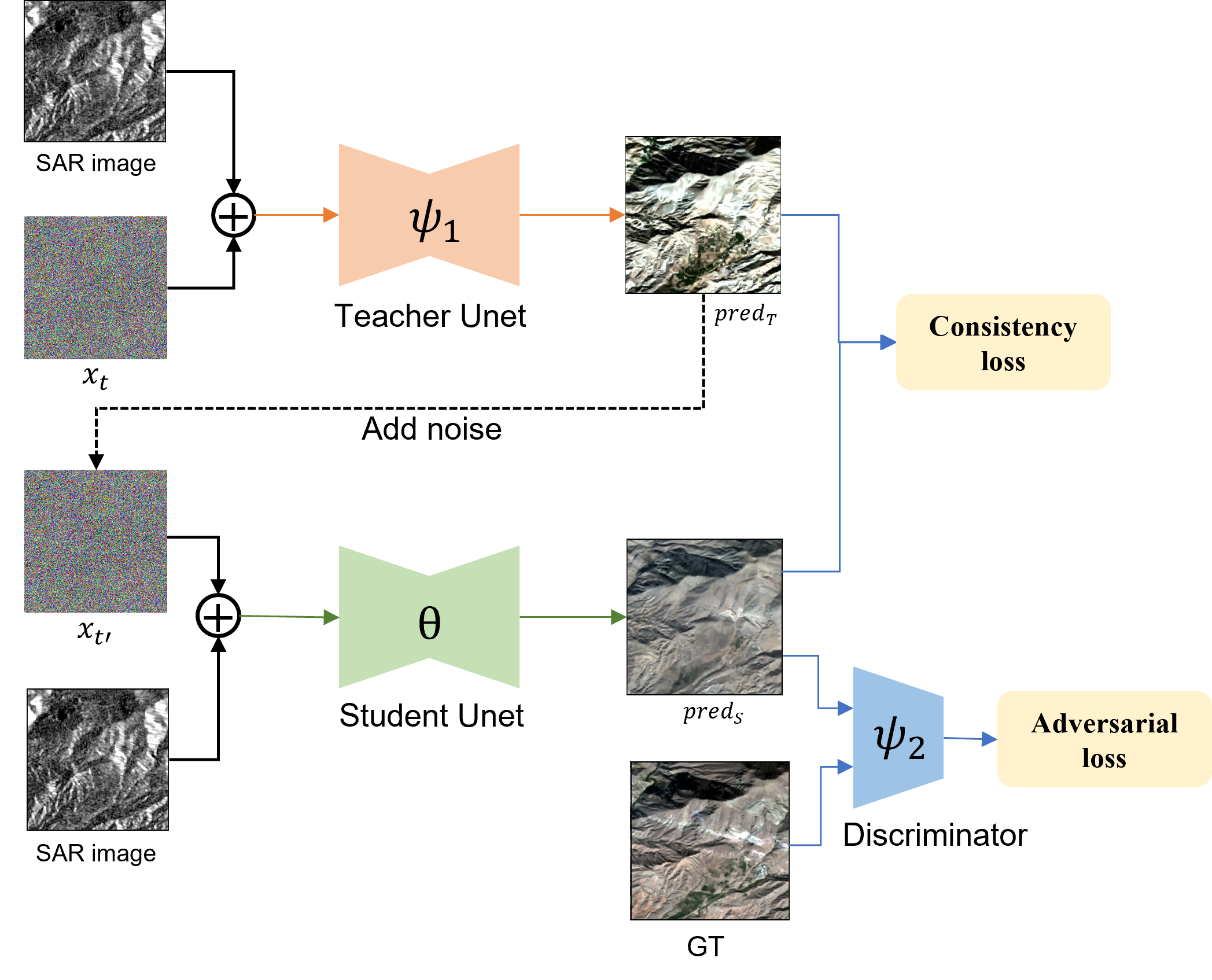}
\caption{The proposed training process involves a student model, a teacher model with fixed weights, and a trainable discriminator. The integration of adversarial learning significantly enhances the quality of SAR-to-optical image translation.}
\label{train_process}
\end{figure}

\subsection{Training Process}
Our proposed training method, as illustrated in Fig.\ref{train_process}, involves three models: a student model, a teacher model with fixed weights, and a trainable discriminator. The rationale behind this design is that the original SAR-to-optical image translation model already serves as an effective feature extractor. By utilizing only the downsampling part, we can effectively extract critical image features while avoiding the computational burden associated with upsampling, thereby enhancing overall training efficiency and stability.

Specifically, we randomly select a time step \( t \) from \([0, T]\). The teacher model receives a noisy optical image \( x_t \) and, conditioned on the SAR image, performs a denoising step to obtain \( x_0 \), denoted as \( \text{pred}_T \). Next, we set \( t' = t - 1 \) and add noise to the optical image generated by the teacher model to create the input for the student model. The student model, along with the SAR image as input, performs a denoising step to produce the output \( \text{pred}_S \).
The generated optical images from the teacher model \( \text{pred}_T \) and the student model \( \text{pred}_S \) are then compared to calculate the consistency loss. This loss reflects the structural and detailed differences between the two outputs, and minimizing this loss guides the student model to match the teacher model's output progressively.
Specifically, we use the Mean Squared Error (MSE) loss, defined as follows:
\begin{equation}
\mathcal{L}_{\text{consistency}} = \frac{1}{N} \sum_{i=1}^{N} \left( \text{pred}_T^{(i)} - \text{pred}_S^{(i)} \right)^2
\end{equation}
where \( N \) is the number of pixels in the image. This MSE loss ensures that the student model's output gradually aligns with the teacher model's output, improving the overall accuracy and quality of the generated optical images.

Subsequently, we input the optical image generated by the student model and the ground truth optical image, both conditioned on the SAR image, into the discriminator. We compute the adversarial loss by comparing the features of the generated and real images. Adversarial learning further enhances the quality of the student model's generated images, making them more closely resemble real optical images. Specifically, we use the hinge loss\cite{lim2017geometric} for the adversarial loss. For the generator \( G \) (student model in our case):
\begin{equation}
\mathcal{L}_{\mathrm{adv}}^{G} = -\mathbb{E}_{\mathbf{x_0}, \mathbf{c_s}} \left[ D_{\psi_2 }(f_{\theta}(\mathbf{x_{t'}}, \mathbf{c_s}), \mathbf{c_s}) \right]
\end{equation}
whereas the discriminator is trained to minimize:
\begin{equation}
\begin{array}{l}
\mathcal{L}_{\mathrm{adv}}^{D} = \mathbb{E}_{\mathbf{x_{0}}}\left[\max \left(0,1-\mathcal{D}_{\psi_2}\left(f_{\theta}\left(\mathbf{x_{0}},\mathbf{c_s}\right),\mathbf{c_s}\right)\right)\right] \\
\quad+\mathbb{E}_{\mathbf{x_t}}\left[\max \left(0,1+\mathcal{D}_{\psi_2}(f_{\theta}(\mathbf{x_{t'}}, \mathbf{c_s}), \mathbf{c_s})\right)\right],
\end{array}
\end{equation}
where \( \mathbf{x_0} \) is the real optical image, \( \mathbf{c_s} \) is the SAR image condition, \( f_{\theta} \) is the function parameterized by the student model, and \( D_{\psi_2} \) is the discriminator. This hinge loss formulation stabilizes the training process and enhances the quality of the images generated by the student model.

The final training loss for the student model combines both consistency and adversarial losses. The consistency loss ensures that the student model's output progressively matches the teacher model's output, while the adversarial loss improves the quality and realism of the generated images. The combined loss function is given by:
\begin{equation}
\mathcal{L}_{\text{total}} = \mathcal{L}_{\text{consistency}} + \lambda_{\text{adv}} \mathcal{L}_{\mathrm{adv}}^{G}
\end{equation}
where \(\lambda_{\text{adv}}\) is a weighting factor that balances the contribution of the adversarial loss relative to the consistency loss. The final loss \(\mathcal{L}_{\text{total}}\) is used to train the student model, ensuring it learns to produce high-quality optical images from SAR inputs with both structural accuracy and visual realism.

\subsection{Sampling Process}
By using a well-trained model \( f_\theta \) obtained through consistency distillation, we can generate optical images with a single denoising step\cite{song2023consistency}. This process involves sampling from the initial Gaussian distribution \( x_T \sim N(0, I) \) and performing a single forward pass of the model. Theoretically, a model trained with consistency distillation can generate optical images in one step. However, the quality of images generated in a single step is typically low. To improve image quality, we can iterate this denoising process by reintroducing noise into the generated optical image and repeating the denoising step, as described in Algorithm \ref{alg:alg1}. 
It is evident that increasing the number of iterations improves image quality and increases inference latency. Therefore, this multi-step sampling process offers flexibility by trading off computational effort for sample quality.
\begin{algorithm}[H]
\caption{Multi-Step Sampling Process}\label{alg:alg1}
\begin{algorithmic}[1]
\REQUIRE Well-trained model $f_{\theta}$, sequence of selected steps $ t_1 > t_2 > \cdots > t_{N-1}$, condition SAR images $c_{s}$.
\STATE $x_{T} \sim \mathcal{N}(\mathbf{0}, \mathbf{I})$
\STATE $x \gets f_{\theta}({x}_T, T)$
\FOR{$n = 1$ to $N-1$}
    \STATE Sample $z \sim \mathcal{N}(0, I)$
    \STATE $\hat{x}_{t_n} \gets x + \sqrt{\frac{t_n}{t_{n-1}} - \epsilon^2} z$
    \STATE $x \gets f_{\theta}(\hat{x}_{t_n}, t_n, c_{s})$
\ENDFOR
\RETURN $x$
\end{algorithmic}
\label{alg1}
\end{algorithm}

\section{Experiment}
\subsection{Datasets and Metrics}
\subsubsection{Datasets}
We primarily evaluated our model on the SEN12 dataset\cite{schmitt2018sen1} to validate its performance. The SEN12 dataset consists of 282,384 pairs of SAR (VV polarization channel) and optical images obtained from Sentinel-1 and Sentinel-2, respectively. The image blocks cover terrestrial areas across various regions and seasons worldwide and have been manually curated to exclude poorly registered image pairs. The SEN12 dataset is a standard benchmark for tasks such as SAR image coloring and SAR image matching. By evaluating our model on this dataset, we can effectively demonstrate its capabilities and advantages using real-world data.

Furthermore, we conducted additional visualization experiments using the GF3 dataset\cite{fu2021reciprocal} further to substantiate the robustness and effectiveness of our approach. The GF3 dataset's SAR data primarily comes from China's GF3 spaceborne SAR, boasting a resolution of 0.51 meters. GF3 is China's first C-band multi-polarization SAR satellite. This dataset encompasses urban and suburban areas, featuring terrain surfaces such as buildings, roads, and vegetation. The optical data were sourced from Google Maps based on geographical coordinates, with a temporal difference of no more than one month from the corresponding SAR images. We ensured precise registration of the image pairs and selected scenes with minimal changes to guarantee accurate correspondence between the SAR and optical images.

To ensure the quality and consistency of our training data, we performed several preprocessing steps on the input images from both the SEN12 and GF3 datasets. Initially, all input images were resized to a consistent resolution to maintain uniformity across the dataset, followed by normalization to a range of [-1, 1] by dividing the pixel values by their maximum possible value. This normalization step stabilized the training process and ensured that the model received input data with consistent scales. To enhance the robustness of our model and prevent overfitting, we applied various data augmentation techniques, including random rotations and horizontal and vertical flips\cite{mumuni2022data}. These augmentations simulated different viewing angles and conditions, thereby improving the model's generalization ability. Finally, the normalized SAR image was concatenated with the optical image along the channel dimension for each training example, allowing the model to learn the relationship between the SAR and optical modalities effectively. These preprocessing steps ensured that the input images were of high quality and suitable for training our SAR-to-optical image translation model, thereby enhancing the overall performance and robustness of our approach.

\subsubsection{Metrics}
To comprehensively evaluate the performance of our model, we used several commonly employed image quality assessment metrics, including the Structural Similarity Index (SSIM)\cite{wang2004image}, Peak Signal-to-Noise Ratio (PSNR)\cite{5596999}, and Fréchet Inception Distance (FID)\cite{heusel2017gans}. 
SSIM measures the similarity between two images in terms of luminance, contrast, and structure. The core idea is that if two images are similar in these aspects, they are also visually similar. SSIM values range from -1 to 1, where 1 indicates that the two images are identical. Unlike traditional mean squared error (MSE), SSIM aligns more closely with human visual perception and is thus widely used for image quality assessment. 
PSNR is another error-based metric used to measure the difference between the reconstructed and original images. PSNR is calculated by taking the logarithm of the ratio of the maximum possible power of the image signal to the power of the distorting noise, expressed in decibels (dB). 
Additionally, we used Fréchet Inception Distance (FID), a metric for evaluating the quality of generated images. FID assesses image quality by comparing the distribution differences between generated and real images in feature space, with lower values indicating that the generated images are closer in quality to real images. FID first uses a pre-trained Inception network to extract image features and then calculates the Fréchet distance between the features of generated images and real images.

By utilizing these metrics, we can comprehensively evaluate the model's performance in the task of translating SAR images to optical images, ensuring high quality in terms of structure, detail, and visual perception. SSIM provides an intuitive measure of image similarity, PSNR offers pixel-level error evaluation, and FID assesses the authenticity and quality of generated images from the perspective of feature distribution. Combining the results from these metrics allows us to gain a thorough understanding of the model's performance.

Additionally, we conducted visualization experiments to further illustrate the effectiveness of our model in translating SAR images to optical images, demonstrating its practical applicability and visual fidelity.

\begin{figure}[htbp]
    \centering
    \begin{tabular}{c}
    \includegraphics[width=0.5\textwidth]{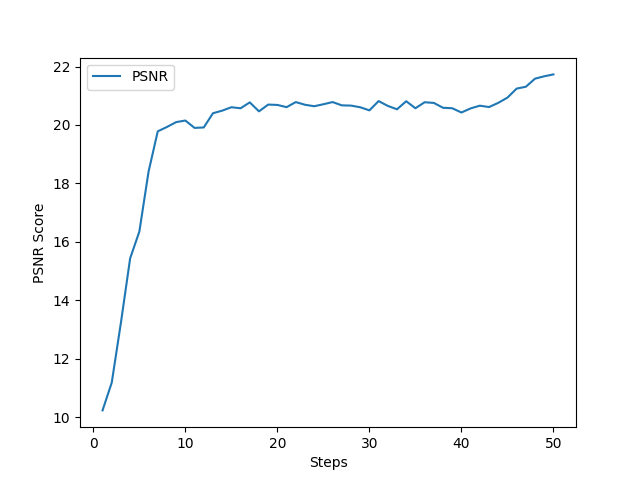}
\\
\ (a) \\
\vspace{1pt} 
    \includegraphics[width=0.5\textwidth]{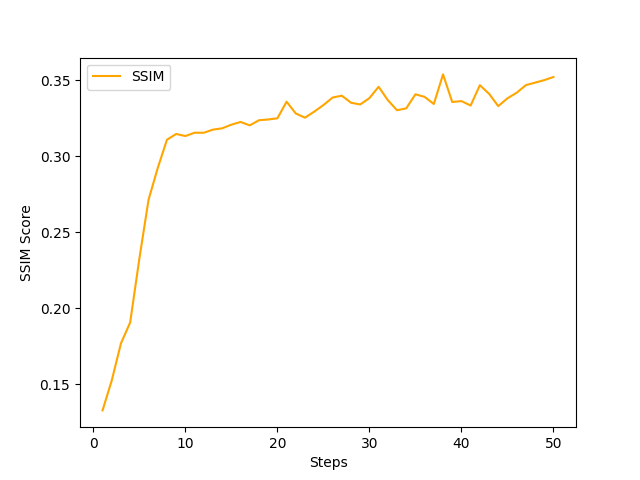}\\
\ (b) \\
    \end{tabular}
    \caption{The relationship between the number of inference iterations and image quality, measured using (a) PSNR and (b) SSIM. The plots show that while low iteration counts result in poor image quality, increasing the iterations improves quality to a point beyond which additional iterations offer minimal benefits.}
    \label{fig_sim}
\end{figure}

\subsection{Implementation Details}
We conducted our model training on a single machine equipped with 4 $\times$ NVIDIA RTX4090 GPUs. 
All three of our models were initialized with the same weights from a pre-trained model with $T=1000$. During the training process, the student and discriminator models were fine-tuned, while the teacher model's weights were kept frozen. 
We utilized the AdamW optimizer\cite{loshchilov2017fixing} with parameters \( \beta_1 = 0.9 \) and \( \beta_2 = 0.999 \). The training process consisted of 50,000 iterations, with the learning rate linearly increased from 0 to \( 8 \times 10^{-6} \) over the first 1,000 iterations and then maintained at \( 8 \times 10^{-6} \) for the remainder of the training. The batch size was set to 16, and the adversarial loss weighting factor \(\lambda_{\text{adv}}\) was set to 0.5.

\subsection{Ablation Study}
In this section, we present an ablation study to evaluate the contributions of various components of our model. This study aims to dissect and understand the impact of key elements, such as the discriminator and the number of inference steps, on the overall performance of our SAR-to-optical image translation framework. By systematically analyzing these components, we can provide a clearer picture of how each part contributes to the quality and efficiency of the generated images, thereby validating the effectiveness of our proposed methods.
\subsubsection{The Effect of the Discriminator}
\begin{figure}[htbp]
    \centering
    \begin{tabular}{@{\hspace{2pt}}c@{\hspace{2pt}}c@{\hspace{2pt}}c@{\hspace{2pt}}c}
    \includegraphics[width=0.115\textwidth]{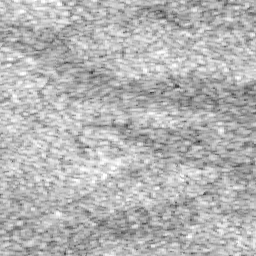}&
    \includegraphics[width=0.115\textwidth]{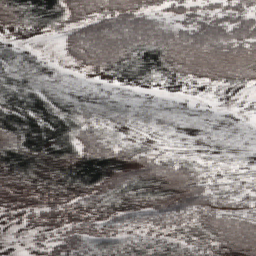}&
    \includegraphics[width=0.115\textwidth]{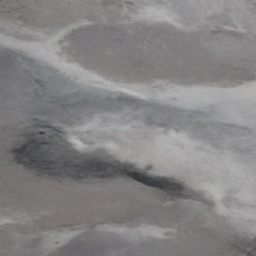}&
    \includegraphics[width=0.115\textwidth]{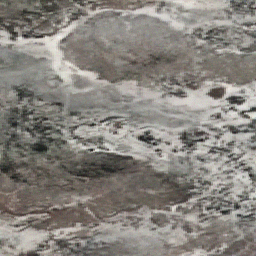}

\\
    \includegraphics[width=0.115\textwidth]{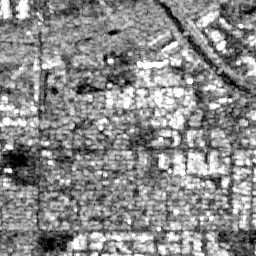}&
    \includegraphics[width=0.115\textwidth]{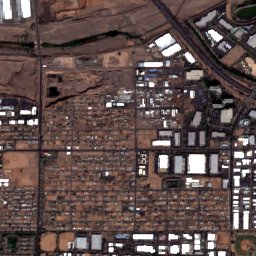}&
    \includegraphics[width=0.115\textwidth]{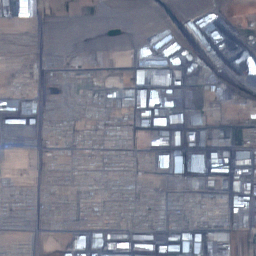}&
    \includegraphics[width=0.115\textwidth]{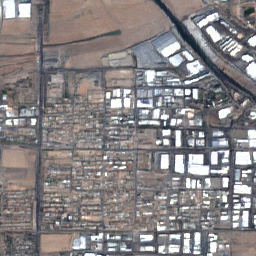}

\\
    \includegraphics[width=0.115\textwidth]{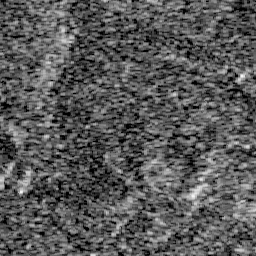}&
    \includegraphics[width=0.115\textwidth]{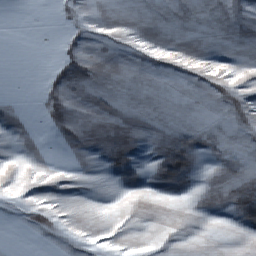}&
    \includegraphics[width=0.115\textwidth]{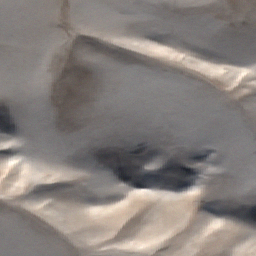}&
    \includegraphics[width=0.115\textwidth]{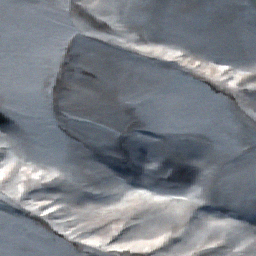}
\\
    \includegraphics[width=0.115\textwidth]{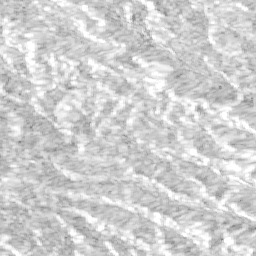}&
    \includegraphics[width=0.115\textwidth]{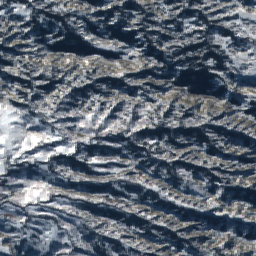}&
    \includegraphics[width=0.115\textwidth]{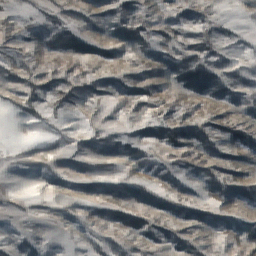}&
    \includegraphics[width=0.115\textwidth]{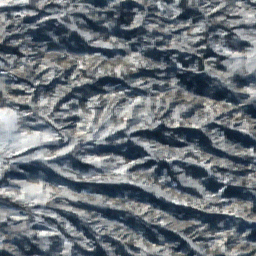}
\\
\ (a) & \ (b) & \ (c) & \ (d) \\
    \end{tabular}
    \caption{Effect of the discriminator on image quality: (a) SAR input images, (b) Ground Truth (GT) optical images, (c) results using basic consistency distillation, and (d) results with adversarial learning enabled. The images in column (c) show blurriness, unclear boundaries, and color shifts, which are significantly mitigated in column (d), demonstrating enhanced quality and clarity due to adversarial learning.}
    \label{fig:no_adv}
\end{figure}

We first demonstrate the effect of the discriminator on the results, as illustrated in Fig.\ref{fig:no_adv},  which are from 8 iterative steps. The third column in Fig.\ref{fig:no_adv} shows the outcomes using basic consistency distillation, while the fourth column displays the results with adversarial learning enabled. It is evident that with only consistency distillation, the images generated at low step counts tend to be blurry, have unclear boundaries, and exhibit overall color shifts. In contrast, these issues are significantly mitigated when adversarial learning is applied. The images produced with adversarial learning are sharper, with clearer boundaries and more accurate color representation. This indicates that including adversarial learning effectively enhances the quality and clarity of the generated optical images, even at low step counts. The comparison clearly shows that adversarial learning is crucial in improving image fidelity and mitigating artifacts common in low-step consistency distillation. Therefore, incorporating adversarial learning into the consistency distillation framework is essential for achieving high-quality optical image translation from SAR images.

\subsubsection{Trade-off Between Inference Speed and Image Quality}
In this section, we present the relationship between the number of inference iterations and image quality scores, specifically using PSNR and SSIM as metrics. As illustrated in Fig.\ref{fig_sim}, when the number of inference steps is deficient (e.g., 1-2 steps), both PSNR and SSIM scores are low, indicating poor image quality. This demonstrates the necessity of iterative inference, where the image is gradually refined by adding noise to the already generated image and then translating it again.
However, as the number of iterations increases, the improvement in scores becomes less pronounced, especially around 16 steps. This suggests that while iterative inference is essential for improving image quality, there is a diminishing return on the benefits of additional iterations beyond a certain point. In other words, increasing the number of iterations improves image quality up to a certain level, after which the gains are minimal. This highlights the trade-off between inference speed and image quality, emphasizing the importance of finding an optimal balance.

\subsection{Experiment Results}
We initially evaluated our model on the SEN12 dataset, conducting both quantitative and qualitative analyses. Due to limited training resources, we have not yet trained and tested our model on large-scale scenarios. To assess the performance of our method compared to existing image translation approaches, we performed comparative analyses with several GAN-based models. 
Specifically, we evaluated our model in comparison to CycleGAN\cite{zhu2017unpaired} and NiceGAN\cite{chen2020reusing}, which have both shown superior performance in image-to-image translation tasks. Additionally, we tested our model against CRAN\cite{fu2021reciprocal}, an innovative adversarial network that utilizes cascaded residual connections and a hybrid L1-GAN loss, specifically tailored for SAR-to-optical image translation. For fairness, we used their official implementations and conducted tests under the same conditions on the same dataset. Furthermore, we compared the results of our model with those of the teacher model S2ODPM\cite{bai2023conditional}. This comprehensive comparison highlights the advantages of our proposed method in generating high-quality optical images from SAR image data, demonstrating its superiority over existing GAN-based models and the teacher model.

\begin{table}[!t]
\centering
\caption{Performance Comparison of different SAR-to-Optical methods.}
\label{tab:mytable}
\begin{tabular}{|c||c||c||c|}
\hline
Method & PSNR$\uparrow$ & FID$\downarrow$ & SSIM$\uparrow$ \\
\hline
CycleGAN & 13.33 & 192.13 & 0.2156  \\
\hline
NiceGAN & 12.48 & 211.62 & 0.2181  \\
\hline
CRAN & 14.13 & 214.46 & 0.2289  \\
\hline
S2ODPM & 18.76 & 136.38 & 0.3020  \\
\hline
\textbf{Ours(8)} & 19.93 & 126.76 & 0.3108 \\
\hline
\textbf{Ours(16)} & \textbf{20.57} & \textbf{116.93} & \textbf{0.3225} \\
\hline
\end{tabular}
\end{table}
\subsubsection{Quantitative Results}
\label{qr}
Table \ref{tab:mytable} presents a comparison of our method with several state-of-the-art techniques using the PSNR, FID, and SSIM metrics on the SEN12 dataset. As shown in Table \ref{tab:mytable}, our model is evaluated with 8 and 16 iterations. Our method consistently outperforms the other techniques across all metrics being assessed. Specifically, our approach achieves the highest PSNR and SSIM scores, as well as the lowest FID score, indicating that the generated images have the highest fidelity, the least amount of noise, and superior quality. This demonstrates that our model effectively captures and reconstructs the fine details of the optical images from SAR inputs, preserving the structural integrity of the original scenes. 
Notably, our model also outperforms the teacher model, highlighting the enhancement in image quality due to adversarial learning and the reduction in cumulative errors from fewer iterations\cite{yang2023diffusion}. Overall, these quantitative results confirm the robustness and superiority of our proposed method in generating high-quality optical images from SAR data, significantly outperforming existing state-of-the-art methods in terms of fidelity, quality, and structural accuracy.
\begin{figure*}[!t]
    \centering
    \begin{tabular}{@{}c@{\hspace{0.5pt}}c@{\hspace{0.5pt}}c@{\hspace{0.5pt}}c@{\hspace{0.5pt}}c@{\hspace{0.5pt}}c@{\hspace{0.5pt}}c@{\hspace{0.5pt}}c@{\hspace{10pt}}}
    \includegraphics[width=0.123\textwidth]{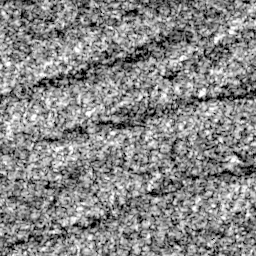}&
    \includegraphics[width=0.123\textwidth]{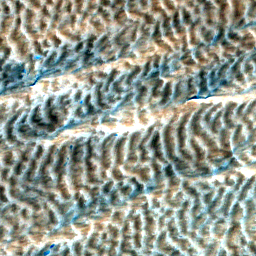}&
    \includegraphics[width=0.123\textwidth]{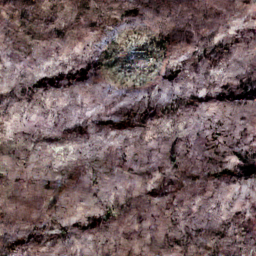}&
    \includegraphics[width=0.123\textwidth]{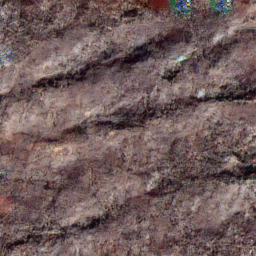}&
    \includegraphics[width=0.123\textwidth]{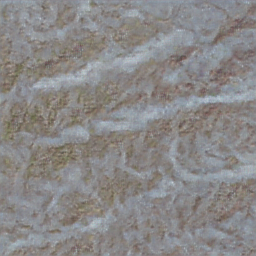}&    
    \includegraphics[width=0.123\textwidth]{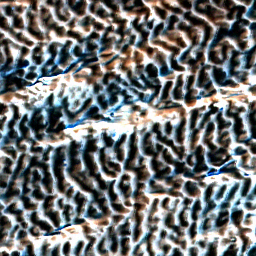}&
    \includegraphics[width=0.123\textwidth]{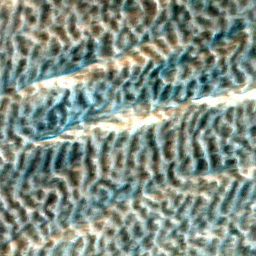}&
    \includegraphics[width=0.123\textwidth]{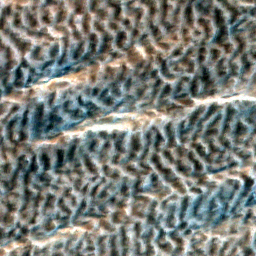}

\\
    \includegraphics[width=0.123\textwidth]{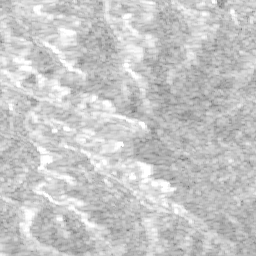}&
    \includegraphics[width=0.123\textwidth]{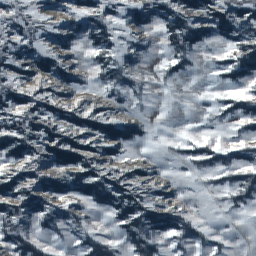}&
    \includegraphics[width=0.123\textwidth]{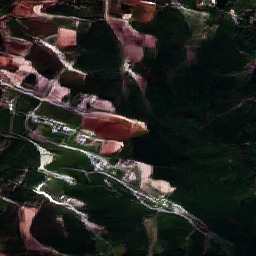}&
    \includegraphics[width=0.123\textwidth]{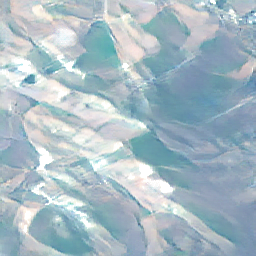}&
    \includegraphics[width=0.123\textwidth]{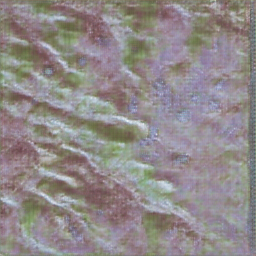}&    
    \includegraphics[width=0.123\textwidth]{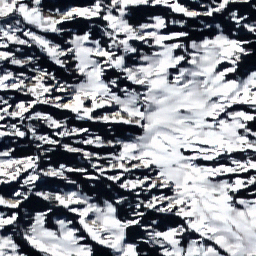}&
    \includegraphics[width=0.123\textwidth]{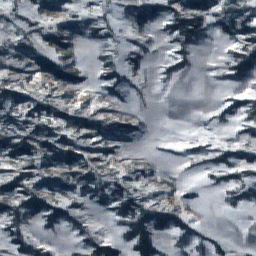}&
    \includegraphics[width=0.123\textwidth]{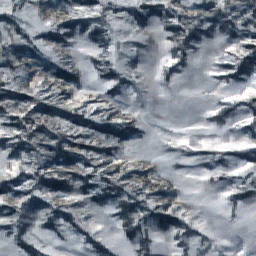}

\\
    \includegraphics[width=0.123\textwidth]{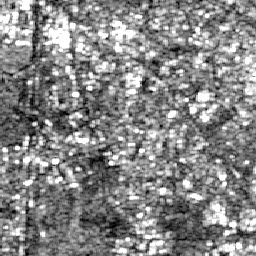}&
    \includegraphics[width=0.123\textwidth]{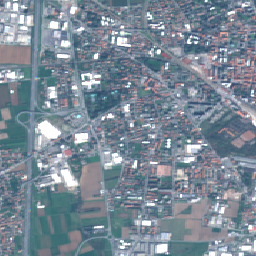}&
    \includegraphics[width=0.123\textwidth]{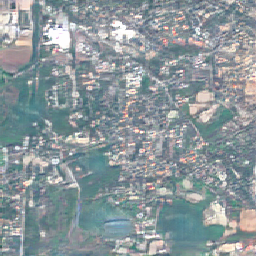}&
    \includegraphics[width=0.123\textwidth]{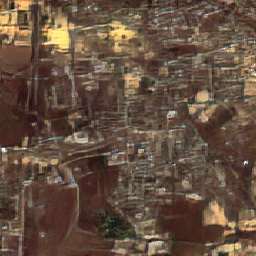}&
    \includegraphics[width=0.123\textwidth]{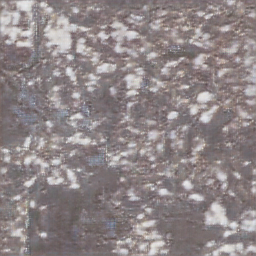}&    
    \includegraphics[width=0.123\textwidth]{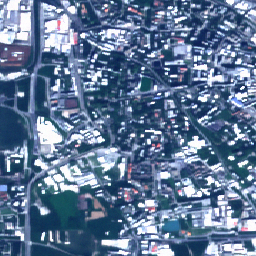}&
    \includegraphics[width=0.123\textwidth]{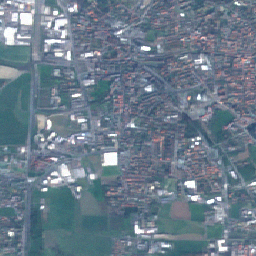}&
    \includegraphics[width=0.123\textwidth]{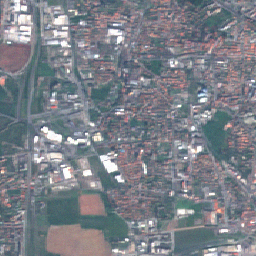}
\\
    \includegraphics[width=0.123\textwidth]{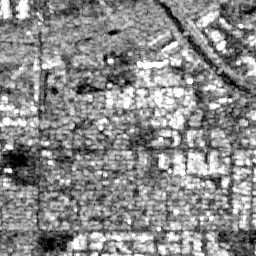}&
    \includegraphics[width=0.123\textwidth]{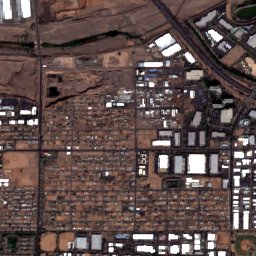}&
    \includegraphics[width=0.123\textwidth]{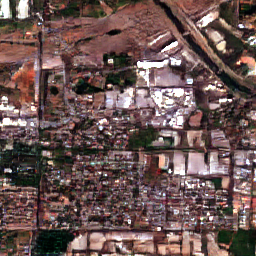}&
    \includegraphics[width=0.123\textwidth]{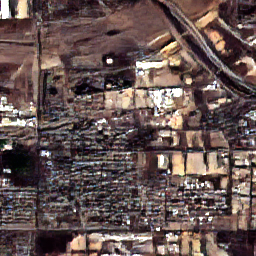}&
    \includegraphics[width=0.123\textwidth]{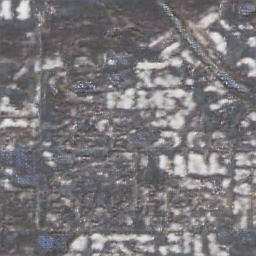}&    
    \includegraphics[width=0.123\textwidth]{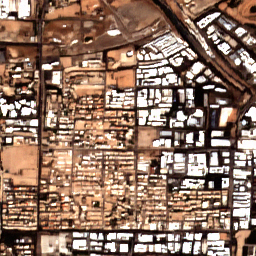}&
    \includegraphics[width=0.123\textwidth]{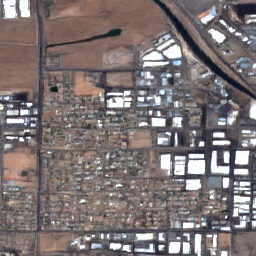}&
    \includegraphics[width=0.123\textwidth]{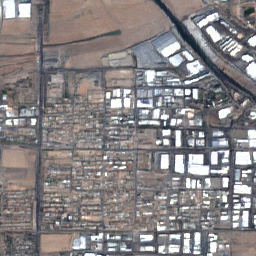}
\\
    \includegraphics[width=0.123\textwidth]{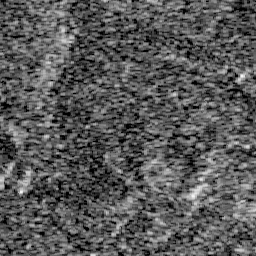}&
    \includegraphics[width=0.123\textwidth]{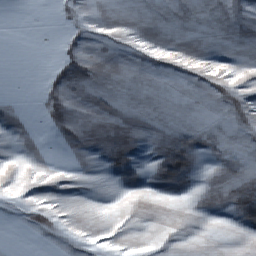}&
    \includegraphics[width=0.123\textwidth]{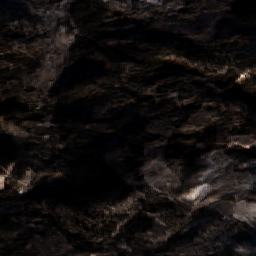}&
    \includegraphics[width=0.123\textwidth]{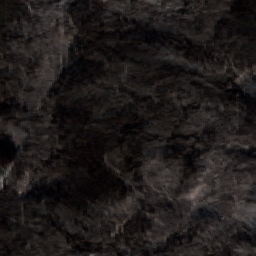}&
    \includegraphics[width=0.123\textwidth]{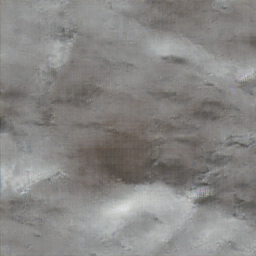}&    
    \includegraphics[width=0.123\textwidth]{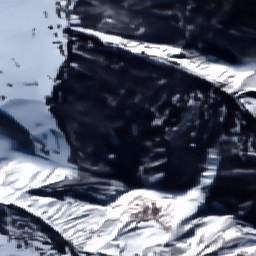}&
    \includegraphics[width=0.123\textwidth]{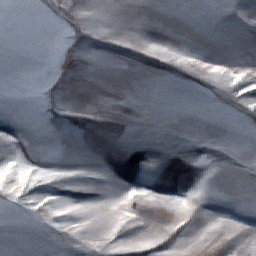}&
    \includegraphics[width=0.123\textwidth]{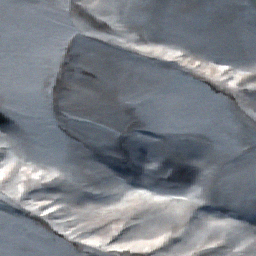}
\\
    \includegraphics[width=0.123\textwidth]{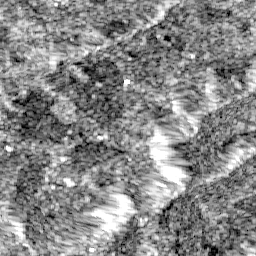}&
    \includegraphics[width=0.123\textwidth]{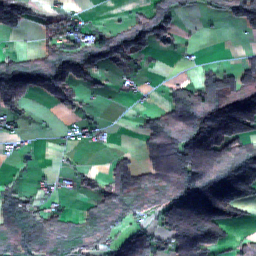}&
    \includegraphics[width=0.123\textwidth]{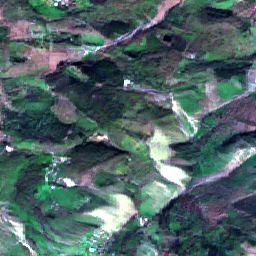}&
    \includegraphics[width=0.123\textwidth]{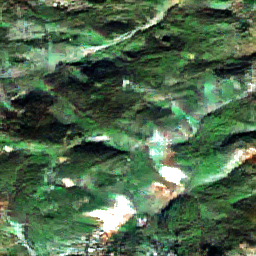}&
    \includegraphics[width=0.123\textwidth]{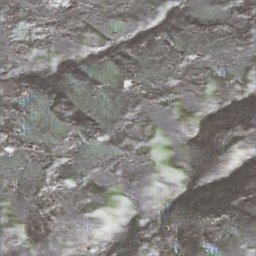}&    
    \includegraphics[width=0.123\textwidth]{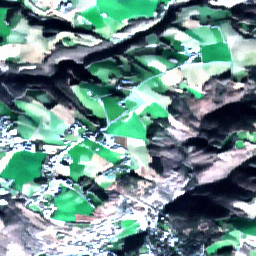}&
    \includegraphics[width=0.123\textwidth]{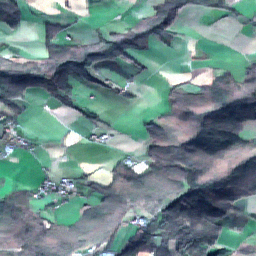}&
    \includegraphics[width=0.123\textwidth]{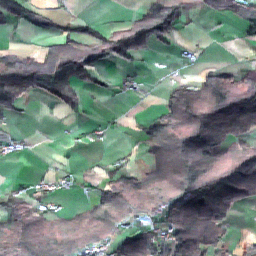}

\\

\ (a)  & \ (b) & \ (c) & \ (d) & \ (e) & \ (f) & \ (g) & \ (h) \\
    \end{tabular}
    \caption{Comparison of SAR-to-optical image translation results on SEN-12 dataset. Each row represents different scenes at various times. Columns from left to right: (a) SAR input images, (b) ground truth (GT) optical images, results from (c) CycleGAN, (d) NiceGAN, (e) CRAN, (f)Teacher model, (g) our model with 16 steps, and (h) our model with 8 steps. The visual comparison demonstrates that our model, particularly at 8 and 16 steps, produces more realistic and natural optical images with clearer boundaries and less blurriness compared to other methods.}
    \label{fig1:q1}
\end{figure*}
\begin{figure*}[!t]
    \centering
    \begin{tabular}{@{}c@{\hspace{0.5pt}}c@{\hspace{0.5pt}}c@{\hspace{0.5pt}}c@{\hspace{0.5pt}}c@{\hspace{0.5pt}}c@{\hspace{0.5pt}}c@{\hspace{0.5pt}}c@{\hspace{10pt}}}
    
    \includegraphics[width=0.123\textwidth]{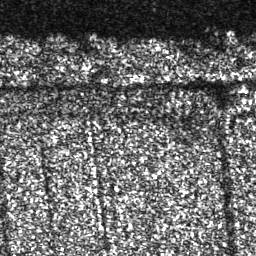}&
    \includegraphics[width=0.123\textwidth]{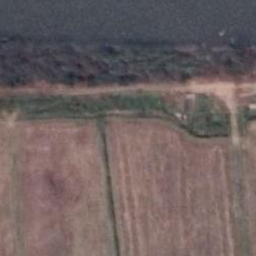}&
    \includegraphics[width=0.123\textwidth]{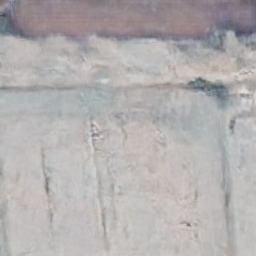}&
    \includegraphics[width=0.123\textwidth]{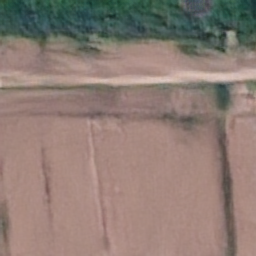}&
    \includegraphics[width=0.123\textwidth]{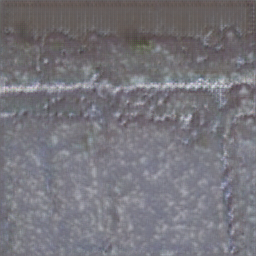}&    
    \includegraphics[width=0.123\textwidth]{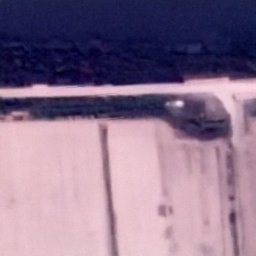}&
    \includegraphics[width=0.123\textwidth]{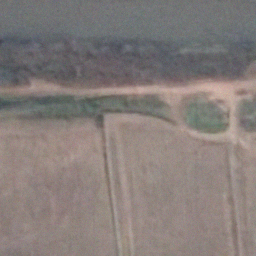}&
    \includegraphics[width=0.123\textwidth]{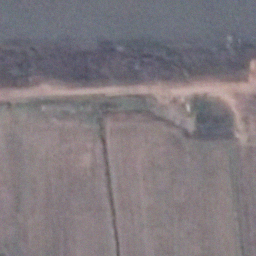}

\\
    \includegraphics[width=0.123\textwidth]{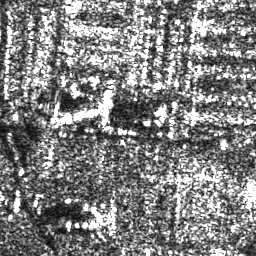}&
    \includegraphics[width=0.123\textwidth]{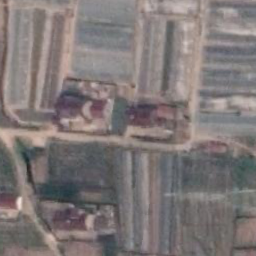}&
    \includegraphics[width=0.123\textwidth]{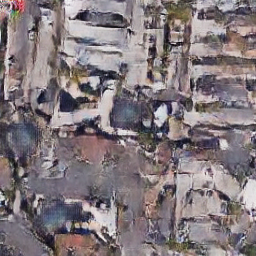}&
    \includegraphics[width=0.123\textwidth]{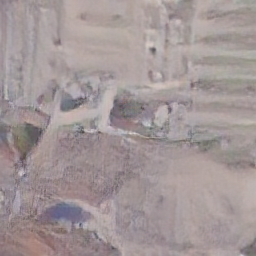}&
    \includegraphics[width=0.123\textwidth]{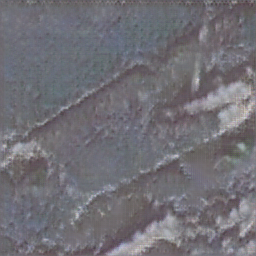}&    
    \includegraphics[width=0.123\textwidth]{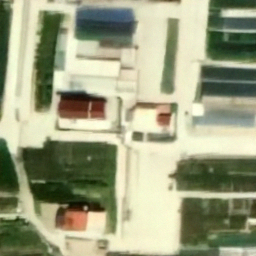}&
    \includegraphics[width=0.123\textwidth]{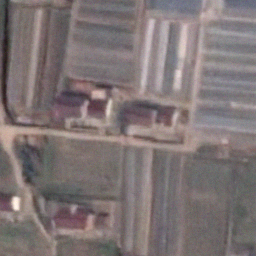}&
    \includegraphics[width=0.123\textwidth]{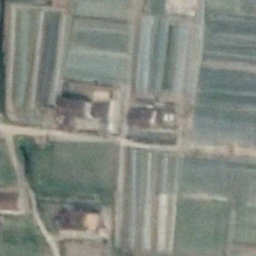}
\\

    \includegraphics[width=0.123\textwidth]{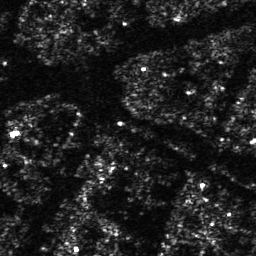}&
    \includegraphics[width=0.123\textwidth]{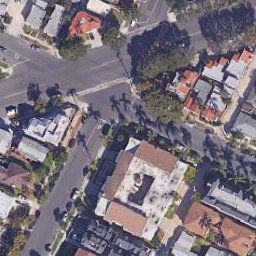}&
    \includegraphics[width=0.123\textwidth]{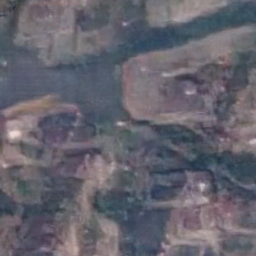}&
    \includegraphics[width=0.123\textwidth]{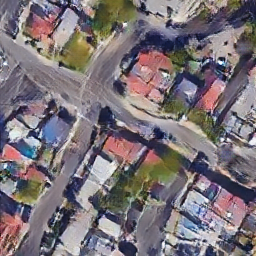}&
    \includegraphics[width=0.123\textwidth]{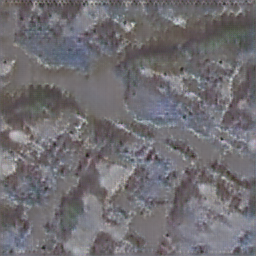}&    
    \includegraphics[width=0.123\textwidth]{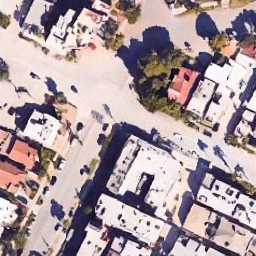}&
    \includegraphics[width=0.123\textwidth]{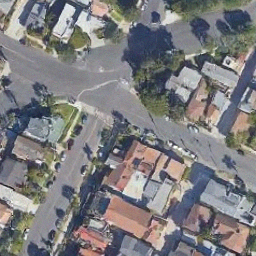}&
    \includegraphics[width=0.123\textwidth]{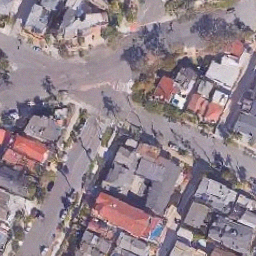}

\\

    \includegraphics[width=0.123\textwidth]{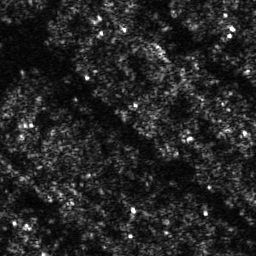}&
    \includegraphics[width=0.123\textwidth]{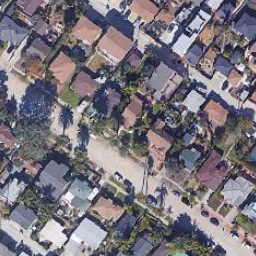}&
    \includegraphics[width=0.123\textwidth]{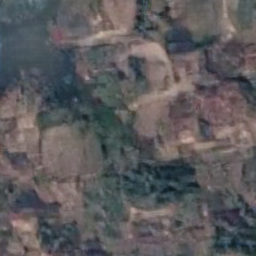}&
    \includegraphics[width=0.123\textwidth]{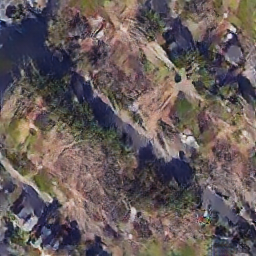}&
    \includegraphics[width=0.123\textwidth]{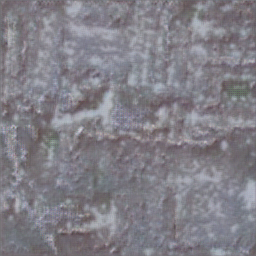}&    
    \includegraphics[width=0.123\textwidth]{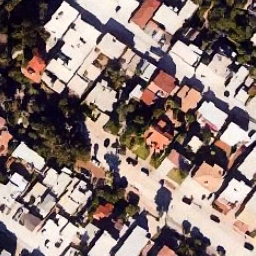}&
    \includegraphics[width=0.123\textwidth]{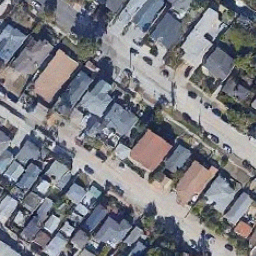}&
    \includegraphics[width=0.123\textwidth]{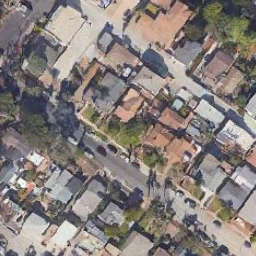}
\\
    \includegraphics[width=0.123\textwidth]{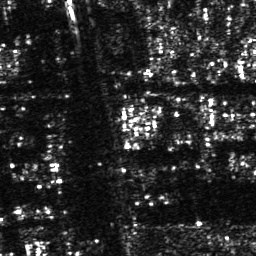}&
    \includegraphics[width=0.123\textwidth]{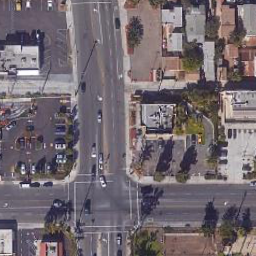}&
    \includegraphics[width=0.123\textwidth]{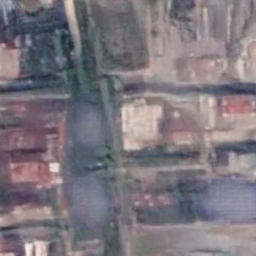}&
    \includegraphics[width=0.123\textwidth]{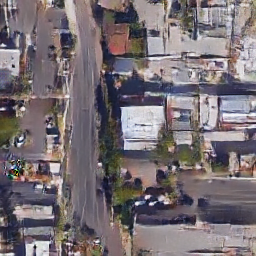}&
    \includegraphics[width=0.123\textwidth]{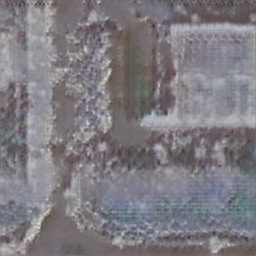}&    
    \includegraphics[width=0.123\textwidth]{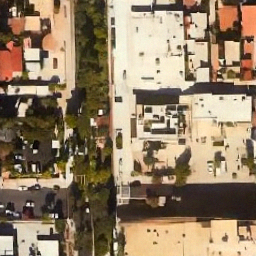}&
    \includegraphics[width=0.123\textwidth]{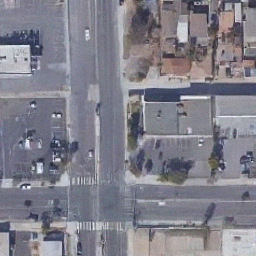}&
    \includegraphics[width=0.123\textwidth]{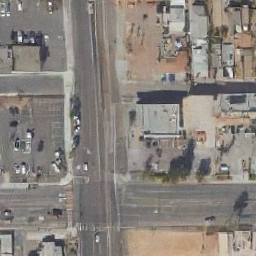}
\\

    \includegraphics[width=0.123\textwidth]{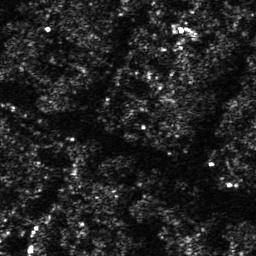}&
    \includegraphics[width=0.123\textwidth]{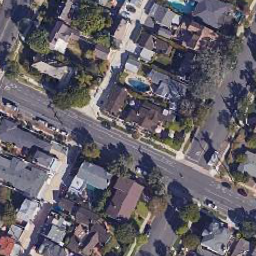}&
    \includegraphics[width=0.123\textwidth]{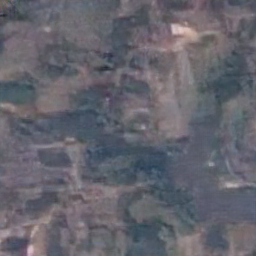}&
    \includegraphics[width=0.123\textwidth]{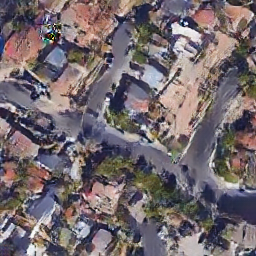}&
    \includegraphics[width=0.123\textwidth]{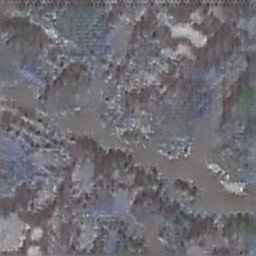}&    
    \includegraphics[width=0.123\textwidth]{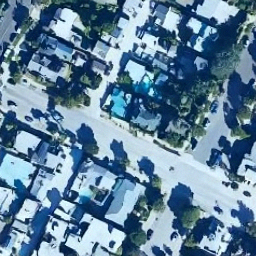}&
    \includegraphics[width=0.123\textwidth]{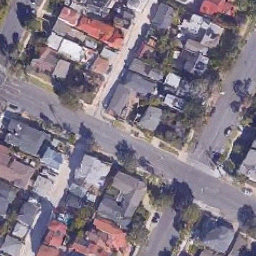}&
    \includegraphics[width=0.123\textwidth]{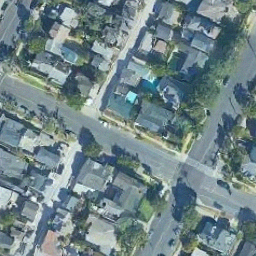}

\\

\ (a)  & \ (b) & \ (c) & \ (d) & \ (e) & \ (f) & \ (g) & \ (h) \\
    \end{tabular}
    \caption{Comparison of SAR-to-optical image translation results on GF3 dataset. Each row represents different scenes at various times. Columns from left to right: (a) SAR input images, (b) ground truth (GT) optical images, results from (c) CycleGAN, (d) NiceGAN, (e) CRAN, (f)Teacher model, (g) our model with 16 steps, and (h) our model with 8 steps.}
    \label{fig1:q2}
\end{figure*}
\begin{table}[!t]
\centering
\caption{Comparison of Inference Latency and Speedup for Different Sampling Methods}
\label{tab:mytable2}
\begin{tabular}{|c||c||c|}
\hline
Method & Latency & Speed Up \\
\hline
 S2ODPM (1000) & 36.75s & 1$\times$ \\
\hline
 DDIM (100) & 3.68s & 10$\times$ \\
\hline
 DPM++ (100) & 3.44s & 11$\times$ \\
 \hline
 Ours (16) & \textbf{0.58s} & 63$\times$ \\
\hline
Ours (8) & \textbf{0.28s} & 131$\times$ \\
\hline
\end{tabular}
\end{table}
\subsubsection{Comparative Analysis of Inference Latency}
To further evaluate the efficiency of our proposed method, we compared the inference latency and speedup ratios of different sampling methods. Specifically, we analyzed the performance of our method against traditional and advanced sampling techniques, including the DDIM\cite{song2020denoising} and DPM++\cite{lu2022dpm} samplers.
The traditional sampling method for the teacher model requires 1000 iterations, while the DDIM and DPM++ samplers use 100 iterations. For our approach, we tested with 16 and 8 iterations of inference.
As shown in Table \ref{tab:mytable2}, traditional sampling methods have the highest inference latency, resulting in the lowest speedup ratio. Specifically, the traditional method (S2ODPM) has a latency of 36.75 seconds with a speedup factor of 1x. In contrast, the DDIM and DPM++ samplers offer substantial improvements, reducing the inference time to 3.68 seconds and 3.44 seconds, respectively, with speedup factors of 10x and 11x.
Our method achieves remarkable inference latencies of just 0.58 seconds and 0.28 seconds for 16 and 8 iterations, respectively, resulting in speedup ratios of 63x and 131x. This significant reduction in inference time demonstrates the effectiveness of our adversarial consistency distillation approach in accelerating the SAR-to-optical image translation process. The enhanced efficiency and reduced latency make our method highly suitable for real-time applications, providing a practical solution for high-quality image generation with minimal computational overhead.

These results show that our approach improves image quality, as evidenced by the superior quantitative scores discussed earlier and significantly reduces inference time. This comparison highlights the flexibility of our method in trading off computational effort for improved sample quality, demonstrating superior performance in both efficiency and effectiveness. Overall, our approach provides a robust and efficient solution for SAR-to-optical image translation, making it ideal for real-time applications in remote sensing.

\subsubsection{Qualitative Results}
In addition to quantitative analysis, we conducted visualization experiments on the test set of the SEN12 dataset to further demonstrate the superiority of our method. Some of the conversion results are shown in Fig.\ref{fig1:q1}. We selected images from various times and scenes to highlight the robustness of our approach.

The visual comparison in Fig.\ref{fig1} highlights the differences between our method and GAN-based methods (CycleGAN, NiceGAN, and CRAN). Compared to these GAN-based models, our method produces more realistic and natural optical images with more explicit boundaries and better color accuracy. The boundaries between objects, such as vegetation, roads, and buildings, are much clearer, indicating a superior ability to preserve fine details. Our method excels in generating images that are closer to the true optical images, capturing intricate textures and structural information that GAN-based methods often miss. The GAN-based approaches tend to produce images with noticeable blurriness and artificial textures, whereas our diffusion model-based method maintains the integrity of the original scene, providing more faithful and high-fidelity translations. The superiority of our method is attributed to several key factors. Firstly, diffusion models, through their iterative denoising process, are less prone to mode collapse and can better handle the complex variations in SAR data. This results in more stable and consistent image generation. Secondly, the integration of adversarial learning in our approach further enhances the quality of the generated images. The adversarial supervision helps refine the details and reduce artifacts, ensuring the generated optical images are more visually accurate and closely resemble the ground truth. Overall, the combination of diffusion models and adversarial learning allows our method to leverage the strengths of both techniques, leading to high-quality image translations that are superior to those produced by traditional GAN-based approaches.

While the teacher model also performs well, it suffers from color deviations and overexposure issues. In contrast, our method demonstrates superior image quality. This improvement is particularly evident in the reduction of blurriness and better preservation of fine details. Despite the fewer inference iterations, our model effectively reduces cumulative errors. Additionally, the incorporation of adversarial supervision enhances the overall image quality. The comparison clearly shows that our model, especially with 8 and 16 iterations, generates optical images that are more visually accurate and closer to the ground truth. This highlights the robustness of our approach in producing high-fidelity images even with reduced computational effort.

Additionally, we compared the results of our method with 8 iterations and 16 iterations. As expected, increasing the number of iterations from 8 to 16 enhances the image quality, further reducing noise and improving the clarity of the translated images. This demonstrates the flexibility of our approach, allowing a trade-off between computational efficiency and image quality.

Overall, these qualitative results underline the advantages of our model in SAR-to-optical image translation, offering a substantial improvement over existing GAN-based methods and even outperforming the teacher model in generating high-fidelity optical images. Additional visualization results, as shown in Fig.\ref{fig1:q2}, were conducted on the GF3 dataset. Compared to the SEN12 dataset, the SAR images in the GF3 dataset contain more speckle noise, posing a greater challenge to the translation capabilities of all models. Despite this, our model performs well, accurately translating the targets with only minor color shifts.

\section{Conclusion}
In this paper, we introduced a novel approach for SAR-to-optical image translation using a diffusion model augmented with adversarial consistency distillation. This method effectively combines the strengths of diffusion models and adversarial learning to generate high-quality optical images with clear boundaries and accurate color representation, even at low iteration counts. Extensive experiments on the SEN12 and GF3 datasets demonstrated that our approach outperforms state-of-the-art GAN-based models in terms of PSNR, SSIM, and FID scores while significantly reducing inference latency, achieving up to a 131-fold speedup. Qualitative results highlight our model's ability to accurately restore fine details and preserve structural integrity, effectively addressing issues like color deviations and overexposure observed in the teacher model. Overall, our method provides a robust and efficient solution for SAR-to-optical image translation, making it highly suitable for real-time remote sensing applications.

Despite the significant advancements achieved, some potential limitations warrant consideration. Firstly, our training process is still demands substantial computational resources, which may limit its accessibility. Integrating adversarial learning, while beneficial for improving image quality, can introduce instability during training, necessitating careful tuning of hyperparameters. Additionally, although our method balances inference speed and image quality effectively, specific applications, especially those requiring ultra-high resolution or real-time processing under constrained computational budgets, may still require further fine-tuning. Addressing these limitations through continued research and development is crucial for enhancing the robustness and applicability of our method. Future work could explore optimizing computational efficiency, improving training stability, and fine-tuning the balance between speed and quality for various applications.


 
\bibliographystyle{IEEEtran}
\bibliography{IEEEabrv, paper}

\begin{thebibliography}{10}
\providecommand{\url}[1]{#1}
\csname url@samestyle\endcsname
\providecommand{\newblock}{\relax}
\providecommand{\bibinfo}[2]{#2}
\providecommand{\BIBentrySTDinterwordspacing}{\spaceskip=0pt\relax}
\providecommand{\BIBentryALTinterwordstretchfactor}{4}
\providecommand{\BIBentryALTinterwordspacing}{\spaceskip=\fontdimen2\font plus
\BIBentryALTinterwordstretchfactor\fontdimen3\font minus \fontdimen4\font\relax}
\providecommand{\BIBforeignlanguage}[2]{{%
\expandafter\ifx\csname l@#1\endcsname\relax
\typeout{** WARNING: IEEEtran.bst: No hyphenation pattern has been}%
\typeout{** loaded for the language `#1'. Using the pattern for}%
\typeout{** the default language instead.}%
\else
\language=\csname l@#1\endcsname
\fi
#2}}
\providecommand{\BIBdecl}{\relax}
\BIBdecl

\bibitem{sun2017recent}
H.~Sun, M.~Shimada, and F.~Xu, ``Recent advances in synthetic aperture radar remote sensing—systems, data processing, and applications,'' \emph{IEEE Geoscience and Remote Sensing Letters}, vol.~14, no.~11, pp. 2013--2016, 2017.

\bibitem{choi2019speckle}
H.~Choi and J.~Jeong, ``Speckle noise reduction technique for sar images using statistical characteristics of speckle noise and discrete wavelet transform,'' \emph{Remote Sensing}, vol.~11, no.~10, p. 1184, 2019.

\bibitem{bai2023conditional}
X.~Bai, X.~Pu, and F.~Xu, ``Conditional diffusion for sar to optical image translation,'' \emph{IEEE Geoscience and Remote Sensing Letters}, 2023.

\bibitem{fu2012research123}
Z.~Fu and W.~Zhang, ``Research on image translation between sar and optical imagery,'' \emph{ISPRS Annals of the Photogrammetry, Remote Sensing and Spatial Information Sciences}, vol.~1, pp. 273--278, 2012.

\bibitem{tiwari2022data}
P.~Tiwari and M.~Ojha, ``Data-centric approach to sar-optical image translation,'' in \emph{International Conference on Computer Vision and Image Processing}.\hskip 1em plus 0.5em minus 0.4em\relax Springer, 2022, pp. 246--260.

\bibitem{zhang2017approach}
W.~ZHANG, G.~TAN, and C.~SUN, ``An approach to translate sar image into optical image,'' \emph{Geomatics and Information Science of Wuhan University}, vol.~42, no.~2, pp. 178--184, 2017.

\bibitem{goodfellow2020generative}
I.~Goodfellow, J.~Pouget-Abadie, M.~Mirza, B.~Xu, D.~Warde-Farley, S.~Ozair, A.~Courville, and Y.~Bengio, ``Generative adversarial networks,'' \emph{Communications of the ACM}, vol.~63, no.~11, pp. 139--144, 2020.

\bibitem{ho2020denoising}
J.~Ho, A.~Jain, and P.~Abbeel, ``Denoising diffusion probabilistic models,'' \emph{Advances in neural information processing systems}, vol.~33, pp. 6840--6851, 2020.

\bibitem{dhariwal2021diffusion}
P.~Dhariwal and A.~Nichol, ``Diffusion models beat gans on image synthesis,'' \emph{Advances in neural information processing systems}, vol.~34, pp. 8780--8794, 2021.

\bibitem{rombach2022high}
R.~Rombach, A.~Blattmann, D.~Lorenz, P.~Esser, and B.~Ommer, ``High-resolution image synthesis with latent diffusion models,'' in \emph{Proceedings of the IEEE/CVF conference on computer vision and pattern recognition}, 2022, pp. 10\,684--10\,695.

\bibitem{cao2024survey}
H.~Cao, C.~Tan, Z.~Gao, Y.~Xu, G.~Chen, P.-A. Heng, and S.~Z. Li, ``A survey on generative diffusion models,'' \emph{IEEE Transactions on Knowledge and Data Engineering}, 2024.

\bibitem{croitoru2023diffusion}
F.-A. Croitoru, V.~Hondru, R.~T. Ionescu, and M.~Shah, ``Diffusion models in vision: A survey,'' \emph{IEEE Transactions on Pattern Analysis and Machine Intelligence}, vol.~45, no.~9, pp. 10\,850--10\,869, 2023.

\bibitem{song2020denoising}
J.~Song, C.~Meng, and S.~Ermon, ``Denoising diffusion implicit models,'' \emph{arXiv preprint arXiv:2010.02502}, 2020.

\bibitem{lu2022dpm}
C.~Lu, Y.~Zhou, F.~Bao, J.~Chen, C.~Li, and J.~Zhu, ``Dpm-solver++: Fast solver for guided sampling of diffusion probabilistic models,'' \emph{arXiv preprint arXiv:2211.01095}, 2022.

\bibitem{song2023consistency}
Y.~Song, P.~Dhariwal, M.~Chen, and I.~Sutskever, ``Consistency models,'' in \emph{International Conference on Machine Learning}.\hskip 1em plus 0.5em minus 0.4em\relax PMLR, 2023, pp. 32\,211--32\,252.

\bibitem{deng2008colorization}
Q.~Deng, Y.~Chen, W.~Zhang, and J.~Yang, ``Colorization for polarimetric sar image based on scattering mechanisms,'' in \emph{2008 Congress on Image and Signal Processing}, vol.~4.\hskip 1em plus 0.5em minus 0.4em\relax IEEE, 2008, pp. 697--701.

\bibitem{uhlmann2013integrating}
S.~Uhlmann and S.~Kiranyaz, ``Integrating color features in polarimetric sar image classification,'' \emph{IEEE Transactions on Geoscience and Remote Sensing}, vol.~52, no.~4, pp. 2197--2216, 2013.

\bibitem{fuentes2019sar}
M.~Fuentes~Reyes, S.~Auer, N.~Merkle, C.~Henry, and M.~Schmitt, ``Sar-to-optical image translation based on conditional generative adversarial networks—optimization, opportunities and limits,'' \emph{Remote Sensing}, vol.~11, no.~17, p. 2067, 2019.

\bibitem{darbaghshahi2021cloud}
F.~N. Darbaghshahi, M.~R. Mohammadi, and M.~Soryani, ``Cloud removal in remote sensing images using generative adversarial networks and sar-to-optical image translation,'' \emph{IEEE Transactions on Geoscience and Remote Sensing}, vol.~60, pp. 1--9, 2021.

\bibitem{yang2022sar}
X.~Yang, J.~Zhao, Z.~Wei, N.~Wang, and X.~Gao, ``Sar-to-optical image translation based on improved cgan,'' \emph{Pattern Recognition}, vol. 121, p. 108208, 2022.

\bibitem{fu2021reciprocal}
S.~Fu, F.~Xu, and Y.-Q. Jin, ``Reciprocal translation between sar and optical remote sensing images with cascaded-residual adversarial networks,'' \emph{Science China Information Sciences}, vol.~64, pp. 1--15, 2021.

\bibitem{wei2023cfrwd}
J.~Wei, H.~Zou, L.~Sun, X.~Cao, S.~He, S.~Liu, and Y.~Zhang, ``Cfrwd-gan for sar-to-optical image translation,'' \emph{Remote Sensing}, vol.~15, no.~10, p. 2547, 2023.

\bibitem{nichol2021improved}
A.~Q. Nichol and P.~Dhariwal, ``Improved denoising diffusion probabilistic models,'' in \emph{International conference on machine learning}.\hskip 1em plus 0.5em minus 0.4em\relax PMLR, 2021, pp. 8162--8171.

\bibitem{songscore}
Y.~Song, J.~Sohl-Dickstein, D.~P. Kingma, A.~Kumar, S.~Ermon, and B.~Poole, ``Score-based generative modeling through stochastic differential equations,'' in \emph{International Conference on Learning Representations}, 2020.

\bibitem{ho2022classifier}
J.~Ho and T.~Salimans, ``Classifier-free diffusion guidance,'' in \emph{NeurIPS 2021 Workshop on Deep Generative Models and Downstream Applications}, 2021.

\bibitem{pooledreamfusion}
B.~Poole, A.~Jain, J.~T. Barron, and B.~Mildenhall, ``Dreamfusion: Text-to-3d using 2d diffusion,'' in \emph{The Eleventh International Conference on Learning Representations}, 2022.

\bibitem{ruiz2023dreambooth}
N.~Ruiz, Y.~Li, V.~Jampani, Y.~Pritch, M.~Rubinstein, and K.~Aberman, ``Dreambooth: Fine tuning text-to-image diffusion models for subject-driven generation,'' in \emph{Proceedings of the IEEE/CVF conference on computer vision and pattern recognition}, 2023, pp. 22\,500--22\,510.

\bibitem{yang2023diffusion}
L.~Yang, Z.~Zhang, Y.~Song, S.~Hong, R.~Xu, Y.~Zhao, W.~Zhang, B.~Cui, and M.-H. Yang, ``Diffusion models: A comprehensive survey of methods and applications,'' \emph{ACM Computing Surveys}, vol.~56, no.~4, pp. 1--39, 2023.

\bibitem{salimans2022progressive}
T.~Salimans and J.~Ho, ``Progressive distillation for fast sampling of diffusion models,'' \emph{arXiv preprint arXiv:2202.00512}, 2022.

\bibitem{meng2023distillation}
C.~Meng, R.~Rombach, R.~Gao, D.~Kingma, S.~Ermon, J.~Ho, and T.~Salimans, ``On distillation of guided diffusion models,'' in \emph{Proceedings of the IEEE/CVF Conference on Computer Vision and Pattern Recognition}, 2023, pp. 14\,297--14\,306.

\bibitem{karras2022elucidating}
T.~Karras, M.~Aittala, T.~Aila, and S.~Laine, ``Elucidating the design space of diffusion-based generative models,'' \emph{Advances in neural information processing systems}, vol.~35, pp. 26\,565--26\,577, 2022.

\bibitem{hyvarinen2009estimation}
A.~Hyv{\"a}rinen, J.~Hurri, P.~O. Hoyer, A.~Hyv{\"a}rinen, J.~Hurri, and P.~O. Hoyer, ``Estimation of non-normalized statistical models,'' \emph{Natural Image Statistics: A Probabilistic Approach to Early Computational Vision}, pp. 419--426, 2009.

\bibitem{song2019generative}
Y.~Song and S.~Ermon, ``Generative modeling by estimating gradients of the data distribution,'' \emph{Advances in neural information processing systems}, vol.~32, 2019.

\bibitem{lu2022dpm1}
C.~Lu, Y.~Zhou, F.~Bao, J.~Chen, C.~Li, and J.~Zhu, ``Dpm-solver: A fast ode solver for diffusion probabilistic model sampling in around 10 steps,'' \emph{Advances in Neural Information Processing Systems}, vol.~35, pp. 5775--5787, 2022.

\bibitem{choi2021ilvr}
J.~Choi, S.~Kim, Y.~Jeong, Y.~Gwon, and S.~Yoon, ``Ilvr: Conditioning method for denoising diffusion probabilistic models,'' in \emph{2021 IEEE/CVF International Conference on Computer Vision (ICCV)}.\hskip 1em plus 0.5em minus 0.4em\relax IEEE, 2021, pp. 14\,347--14\,356.

\bibitem{saharia2022palette}
C.~Saharia, W.~Chan, H.~Chang, C.~Lee, J.~Ho, T.~Salimans, D.~Fleet, and M.~Norouzi, ``Palette: Image-to-image diffusion models,'' in \emph{ACM SIGGRAPH 2022 conference proceedings}, 2022, pp. 1--10.

\bibitem{ronneberger2015u}
O.~Ronneberger, P.~Fischer, and T.~Brox, ``U-net: Convolutional networks for biomedical image segmentation,'' in \emph{Medical image computing and computer-assisted intervention--MICCAI 2015: 18th international conference, Munich, Germany, October 5-9, 2015, proceedings, part III 18}.\hskip 1em plus 0.5em minus 0.4em\relax Springer, 2015, pp. 234--241.

\bibitem{sauer2023adversarial}
A.~Sauer, D.~Lorenz, A.~Blattmann, and R.~Rombach, ``Adversarial diffusion distillation,'' \emph{arXiv preprint arXiv:2311.17042}, 2023.

\bibitem{lim2017geometric}
J.~H. Lim and J.~C. Ye, ``Geometric gan,'' \emph{arXiv preprint arXiv:1705.02894}, 2017.

\bibitem{schmitt2018sen1}
M.~Schmitt, L.~H. Hughes, and X.~X. Zhu, ``The sen1-2 dataset for deep learning in sar-optical data fusion,'' \emph{arXiv preprint arXiv:1807.01569}, 2018.

\bibitem{mumuni2022data}
A.~Mumuni and F.~Mumuni, ``Data augmentation: A comprehensive survey of modern approaches,'' \emph{Array}, vol.~16, p. 100258, 2022.

\bibitem{wang2004image}
Z.~Wang, A.~C. Bovik, H.~R. Sheikh, and E.~P. Simoncelli, ``Image quality assessment: from error visibility to structural similarity,'' \emph{IEEE transactions on image processing}, vol.~13, no.~4, pp. 600--612, 2004.

\bibitem{5596999}
A.~Horé and D.~Ziou, ``Image quality metrics: Psnr vs. ssim,'' in \emph{2010 20th International Conference on Pattern Recognition}, 2010, pp. 2366--2369.

\bibitem{heusel2017gans}
M.~Heusel, H.~Ramsauer, T.~Unterthiner, B.~Nessler, and S.~Hochreiter, ``Gans trained by a two time-scale update rule converge to a local nash equilibrium,'' \emph{Advances in neural information processing systems}, vol.~30, 2017.

\bibitem{loshchilov2017fixing}
\BIBentryALTinterwordspacing
I.~Loshchilov and F.~Hutter, ``Decoupled weight decay regularization,'' 2019. [Online]. Available: \url{https://arxiv.org/abs/1711.05101}
\BIBentrySTDinterwordspacing

\bibitem{zhu2017unpaired}
J.-Y. Zhu, T.~Park, P.~Isola, and A.~A. Efros, ``Unpaired image-to-image translation using cycle-consistent adversarial networks,'' in \emph{Proceedings of the IEEE international conference on computer vision}, 2017, pp. 2223--2232.

\bibitem{chen2020reusing}
R.~Chen, W.~Huang, B.~Huang, F.~Sun, and B.~Fang, ``Reusing discriminators for encoding: Towards unsupervised image-to-image translation,'' in \emph{Proceedings of the IEEE/CVF conference on computer vision and pattern recognition}, 2020, pp. 8168--8177.

\end{thebibliography}
%

 




\vfill

\end{document}